\documentclass[11pt,a4paper]{lumia}
\usepackage[sort&compress]{natbib}
\usepackage{graphicx}
\usepackage{mathtools}
\usepackage{algorithm}
\usepackage{algpseudocode}
\usepackage{array}
\usepackage{multirow}
\usepackage{makecell}
\usepackage{wrapfig}
\usepackage{placeins}
\definecolor{unigroundteal}{HTML}{005451}
\definecolor{unigroundlink}{HTML}{239BA7}
\definecolor{unigroundsoft}{HTML}{EDF6F6}
\definecolor{unigroundrow}{HTML}{E3F1F1}
\definecolor{unigroundgroup}{HTML}{D5E9E8}
\colorlet{lumiateal}{unigroundteal}
\colorlet{lumialink}{unigroundlink}
\colorlet{lumiaabsbg}{unigroundsoft}
\hypersetup{
  colorlinks=true,
  linkcolor=unigroundteal,
  citecolor=unigroundlink,
  urlcolor=unigroundlink,
  pdftitle={UniGround: Universal 3D Visual Grounding via Training-Free Scene Parsing},
  pdfauthor={Jiaxi Zhang et al.}
}

\newcommand{\best}[1]{\textbf{\textcolor{unigroundteal}{#1}}}

\setheadertext{%
  \raisebox{-0.05\height}{\includegraphics[height=36pt,keepaspectratio]{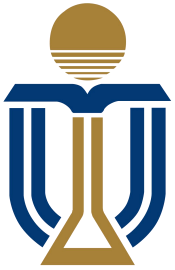}}%
}
\setheadertitle{UniGround: Universal 3D Visual Grounding}
\renewcommand{\today}{August 2026}

\title{UniGround: Universal 3D Visual Grounding via Training-Free Scene Parsing}

\author{%
  {\normalfont\bfseries\fontsize{10}{12}\selectfont
    Jiaxi Zhang\textsuperscript{1,*},
    Yunheng Wang\textsuperscript{1,*},
    Wei Lu\textsuperscript{1},
    Taowen Wang\textsuperscript{1},
    Shuning Zhang\textsuperscript{1},
    Yixiao Feng\textsuperscript{1},\par
    Junzhe Xu\textsuperscript{1,2},
    Shunwang Sun\textsuperscript{3},
    Weisheng Xu\textsuperscript{1},
    Yuetong Fang\textsuperscript{1,\textdagger},
    Renjing Xu\textsuperscript{1,\textdagger}\par}
  \vspace{3pt}%
  {\normalfont\fontsize{9.5}{12}\selectfont
    \textsuperscript{1}The Hong Kong University of Science and Technology (Guangzhou)\\
    \textsuperscript{2}JD Explore Academy
    \qquad
    \textsuperscript{3}Zhejiang University\par}
}

\correspondingemail{%
  \textsuperscript{*}Equal contribution
  \qquad
  \textsuperscript{\textdagger}Corresponding authors
}

\begin{document}

\begin{abstract}

3D Visual Grounding (3DVG) localizes objects from natural-language descriptions in 3D scenes and is fundamental to embodied AI applications. Although foundation models enable open-vocabulary reasoning, they typically rely on pre-generated candidates, creating two sequential bottlenecks. The \emph{candidate bottleneck} occurs when dataset-specific 3D proposal models miss, fragment, or incorrectly group targets under distribution shifts, excluding them from VLM reasoning. The \emph{evidence bottleneck} stems from incomplete visual evidence: global renderings preserve spatial context but obscure object details, whereas candidate-centric views capture local appearance but lack global context. To address these bottlenecks, we propose UniGround, a zero-shot 3DVG framework that addresses both bottlenecks through Global Candidate Filtering and Contextual Precision Grounding. Global Candidate Filtering constructs topology-consistent, class-agnostic candidates from 3D topology and multi-view semantic cues, without dataset-trained 3D detectors, task-specific proposal supervision, or predefined box and category priors. Contextual Precision Grounding jointly reasons over global spatial context and candidate-centric visual evidence, followed by closed-loop consistency verification for reliable target identification. UniGround achieves 46.1\%/34.1\% Acc@0.25/0.5 on ScanRefer and 28.7\% Acc@0.25 on the evaluated ARKitScenes subset of EmbodiedScan. Further experiments demonstrate competitive grounding without dataset-specific 3D priors, cross-dataset generalization to unseen indoor scenes, and robustness to real-world reconstruction noise and practical domain shifts.

\end{abstract}

\maketitle

\section{Introduction}

% \begin{figure*}[!t]
%     \centering
%     \includegraphics[width=\textwidth]{figure/intro.png}
%     \caption{
%     Conceptual comparison and overview of UniGround.
%     \textbf{Top:} Existing zero-shot 3D visual grounding pipelines typically construct candidate objects using dataset-trained 3D detectors with prior information, which may degrade under domain shift and affect downstream reasoning quality.
%     \textbf{Bottom:} UniGround adopts a training-free paradigm consisting of \emph{Global Candidate Filtering} for 2D--3D instance construction and open-vocabulary encoding, followed by the \emph{Local Precision Grounding} for geometric-aware multi-view reasoning with a structured and parsable output.
%     \textbf{Right:} Example of real-world deployment in an unseen office scene.
%     }
%     \label{fig:intro}
% \end{figure*}

3D Visual Grounding (3DVG) is a cornerstone of embodied intelligence that bridges human communication and machine spatial understanding. Specifically, a 3DVG system takes a natural language description as input and precisely locates the corresponding target object within a 3D scene, enabling a wide range of real-world applications, including augmented reality~\cite{liu2021deep,liu2023raydf,liu2025unleashing}, vision-language navigation~\cite{zhang2024navid,wang2025dreamnav,zhou2024navgpt}, and robotic perception~\cite{kong2023rethinking,kong2023robo3d,li2022coarse3d}. Recently, with the rapid rise of large-scale pretrained foundation models, open-vocabulary 3DVG systems are now capable of zero-shot semantic reasoning and flexible natural language instruction following, yet remain largely constrained to in-distribution 3D environments seen during training. This limitation hinders reliable deployment in open-world environments.

\begin{figure*}[t]
    \centering
    \includegraphics[width=\textwidth]{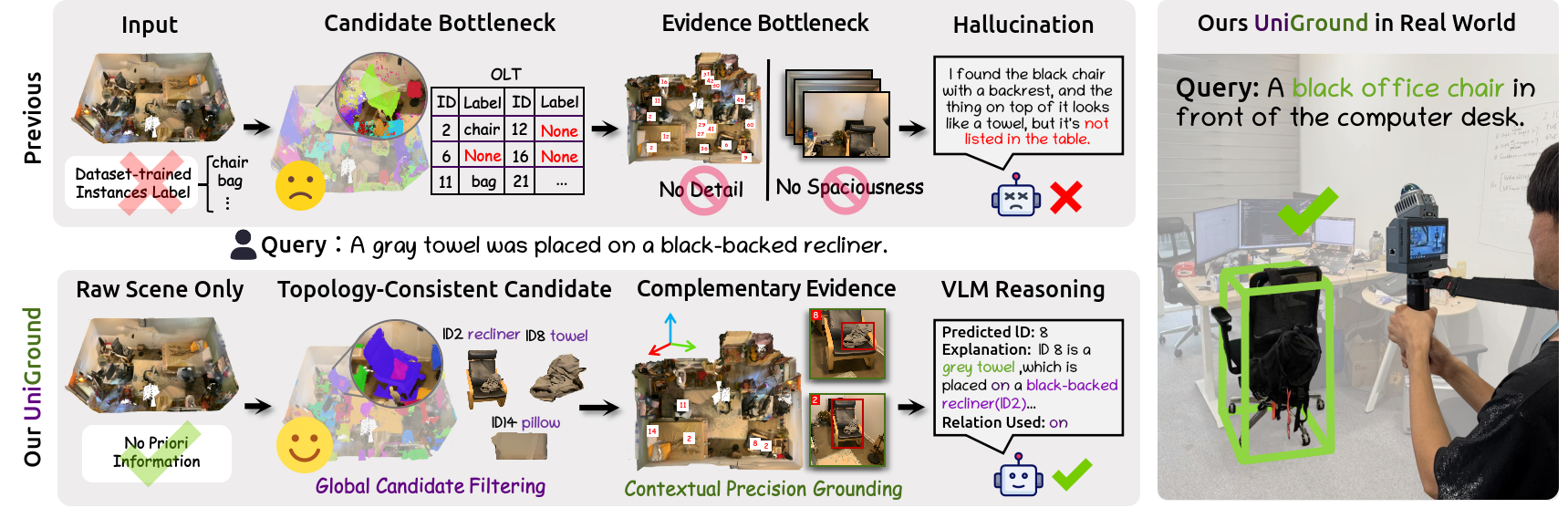}
    \caption{Overview of UniGround. \textbf{Top left:} Prior methods depend on dataset-specific proposals and limited visual evidence. \textbf{Bottom left:} UniGround combines training-free candidate filtering with contextual precision grounding. \textbf{Right:} Real-world deployment.}
    \label{fig:intro}
\end{figure*}

% Existing approaches delegate scene parsing to supervised 3D instance segmentation or detection models, whose object proposals serve as the perceptual input to the foundation model (Fig.~\ref{fig:intro}). As these 3D perception modules are trained on closed-set, domain-specific datasets, they inherently restrict which objects and spatial configurations can be surfaced as candidates, regardless of how semantically capable the downstream model is. In other words, the open-vocabulary capability of such systems is limited to the language reasoning side, while the 3D perception front-end remains fundamentally closed-set. Therefore, 3DVG systems relying on such pretrained 3D modules face two compounding limitations: ~\ding{182} {Limited Generalization}: Reliance on supervised 3D segmentation models trained on in-domain data limits the system's ability to generalize to open-set object categories and unseen environments, rendering any out-of-distribution object undetectable to the grounding pipeline; ~\ding{183} {Poor Comprehension}: Single-perspective visual prompting provides impoverished spatial and semantic context to the VLM, leaving fine-grained object attributes and global scene-level relationships insufficiently grounded, ultimately undermining localization accuracy.

Existing approaches rely on supervised 3D segmentation or detection models to generate object proposals for foundation models (Fig.~\ref{fig:intro}). Trained on closed-set, domain-specific data, these perception front-ends constrain accessible objects and spatial configurations irrespective of downstream semantic capabilities, making such systems open-vocabulary in language reasoning but closed-set in 3D perception. Consequently, they face two compounding limitations: ~\ding{182} {Limited Generalization}: Reliance on supervised 3D segmentation models trained on in-domain data limits the system's ability to generalize to open-set object categories and unseen environments, rendering any out-of-distribution object undetectable to the grounding pipeline; ~\ding{183} {Poor Comprehension}: Single-perspective visual prompting provides impoverished spatial and semantic context to the VLM, leaving fine-grained object attributes and global scene-level relationships insufficiently grounded, ultimately undermining localization accuracy.

%Single-perspective prompting often provides incomplete context, which weakens VLM comprehension and destabilizes localization by missing fine-grained semantics or global spatial relations.

%As shown in Fig.~\ref{fig:intro}, Current mainstream zero-shot 3DVG approaches generally follow a two-stage process. Initially, they leverage prior scene knowledge to drive a supervised 3D segmentation model in constructing a comprehensive Object Lookup Table, which populates the scene with candidate bounding boxes and their associated labels. Then, 3DVG reduces to cross-modal matching within the predefined table via single-perspective prompts, using either global 2D scene projections or per-candidate multi-view RGB renderings

We argue that the bottleneck lies, instead, in how well the model sees. 3D scene understanding can be decomposed into geometric perception, which captures spatial structure and inter-object relationships, and semantic reasoning, which aligns natural language descriptions with visual concepts. While many existing works aim to solve these two problems at once using a larger and more powerful supervised 3D segmentation model, this demands expensive 3D annotations and constraining geometric perception within the semantic biases inherited from pretraining. Counterintuitively, we demonstrate that moving away from domain-specific 3D models is precisely the key to unlocking full-scene generalization in 3DVG. Motivated by this, we propose UniGround, a framework that fundamentally departs from the prevailing paradigm by decoupling geometric perception from semantic reasoning into two distinct stages. UniGround constructs a topology-aware scene representation by integrating 2D segmentation with spatial topology graphs and multi-view RGB information, requiring no 3D supervision of any kind. 
A training-free reasoning framework then grounds this representation through multi-dimensional context prompting, enabling the VLM to perform precise spatial reasoning. Empirically, UniGround attains 46.1/34.1 Acc@0.25/0.5 on ScanRefer and transfers to EmbodiedScan with 28.7 Acc@0.25, validating training-free open-world grounding across unseen scenes. Our main contributions are summarized as follows:

\begin{itemize}
    \item We propose \textbf{Global Candidate Filtering}, which constructs topology-consistent, class-agnostic candidates using 3D topology and multi-view semantic cues, without 3D supervision or predefined box and category priors.

    \item We propose \textbf{Contextual Precision Grounding}, which integrates global spatial context with candidate-centric visual evidence and applies closed-loop semantic--spatial consistency verification for reliable target identification.

    \item We evaluate UniGround on ScanRefer, EmbodiedScan, and real-world scenes, demonstrating competitive zero-shot grounding, cross-dataset generalization, and robustness to reconstruction noise and practical domain shifts.
\end{itemize}

% direction
% LERF: Language Embedded Radiance Fields
% OpenScene: 3D Scene Understanding with Open Vocabularies
% Optimal Scene Graph Planning with Large Language Model Guidance
% Zoo3D: Zero-Shot 3D Object Detection at Scene Level

\section{Related Work}

\subsection{Supervised 3D Visual Grounding}
Early supervised 3D visual grounding methods, exemplified by ScanRefer~\cite{chen2020scanrefer} and ReferIt3D~\cite{achlioptas2020referit3d}, predominantly followed a decoupled ``detect--match'' paradigm~\cite{achlioptas2020referit3d,chen2020scanrefer,jain2022bottom,li2025cityanchor,zhang2023multi3drefer,zhu20233dvista}, using a pretrained 3D detector to generate object proposals and a cross-modal model to match them with language queries. Despite its conceptual simplicity, this paradigm requires costly 3D annotations and inherits the detector's sensitivity to variations in point-cloud density and scene distribution. Subsequent end-to-end approaches jointly learn geometric and linguistic representations for deeper multimodal reasoning~\cite{guo2025tsp3d,huang2025viewsrd,huang20253d,lin2025groundflow,qi2025gpt4scene,qian2024mcln,unal2024concretenet,G3_LQ}, yet their geometric perception remains tied to the point-cloud statistics, object categories, and spatial layouts observed during training. Recent open-vocabulary methods leverage LLM/VLM priors to reduce annotation dependence~\cite{SeeGround,View_on_Graph,VLM_Grounder,SeqVLM,SPAZER,huang2025openground}, but still rely on pretrained 3D components, yielding zero-shot semantic reasoning without comparable geometric generalization. In contrast, UniGround replaces domain-specific 3D models with training-free geometric perception, enabling precise semantic--spatial grounding and robust generalization across unseen scenes.

% 3DVGs
% SeeGround: See and Ground for Zero-Shot Open-Vocabulary 3D Visual Grounding
% View-on-Graph: Zero-shot 3D Visual Grounding via Vision-Language Reasoning on Scene Graphs
% VLM-Grounder: A VLM Agent for Zero-Shot 3D Visual Grounding
% SeqVLM: Proposal-Guided Multi-View Sequences Reasoning via VLM for Zero-Shot 3D Visual Grounding
% SPAZER: Spatial-Semantic Progressive Reasoning Agent for Zero-shot 3D Visual Grounding
% OpenGround: Active Cognition-based Reasoning for Open-World 3D Visual Grounding

% More recent research has shifted toward unified end-to-end architectures that jointly learn 3D geometric and linguistic representations~\cite{}, enabling deeper multimodal fusion so that language cues can directly guide feature extraction and spatial reasoning for more context-aware localization. Although supervised 3DVG achieves strong benchmark results, its dependence on expensive 3D annotations limits scalability and practicality in open-world settings. To alleviate this, recent zero-shot methods exploit LLM/VLM priors for annotation-free grounding~\cite{}. Along this line, we propose UniGround, which eliminates additional labeling while enhancing cross-scene generalization and semantic–spatial comprehension, leading to more accurate and robust localization.

\begin{figure*}[!t]
    \centering
    \includegraphics[width=\textwidth]{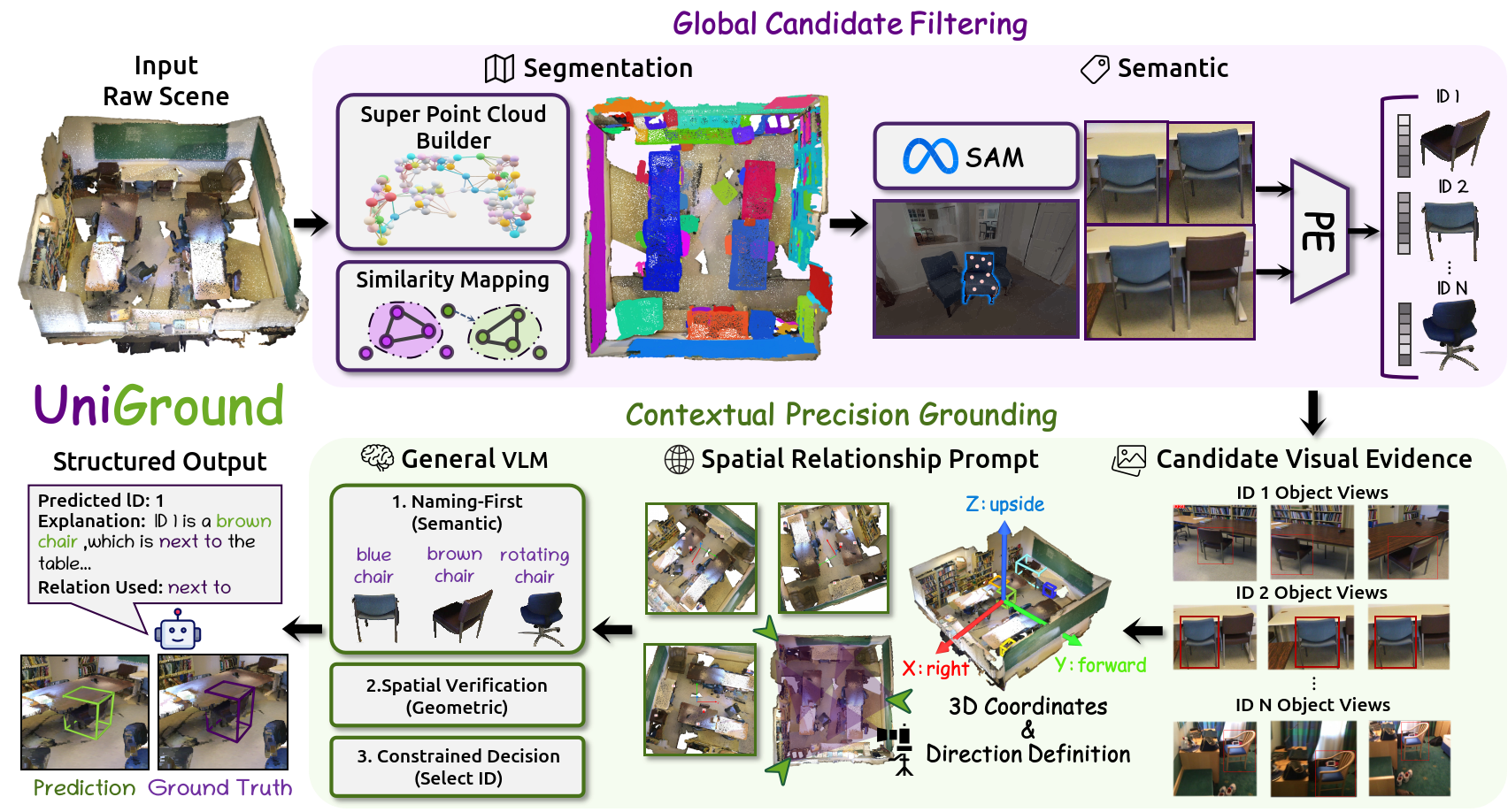}
    \caption{Overview of UniGround. Global Candidate Filtering constructs topology-consistent, query-relevant candidates without supervised 3D proposals. Contextual Precision Grounding identifies the referent by integrating global context, candidate-centric evidence, and closed-loop consistency verification.}
    \label{fig:pipeline}
\end{figure*}

\subsection{Zero-shot 3D Visual Grounding} 
Zero-shot 3DVG aims to leverage the reasoning and cross-modal alignment capabilities of foundation models to overcome the limited generalization of supervised methods.  However, the majority of existing approaches~\cite{SeeGround,View_on_Graph,VLM_Grounder,SeqVLM,SPAZER,huang2025openground} remain zero-shot only in the linguistic sense: they rely on a pretrained 3D instance segmentation model to construct an object lookup table that supplies bounding boxes and semantic labels, reducing grounding to a filtering problem over predefined object candidates. This design enables open-vocabulary query handling, but geometric perception remains anchored to in-domain 3D supervision, meaning these methods generalize across language, but not across scenes. For instance, SeeGround~\cite{SeeGround} and View-on-Graph~\cite{View_on_Graph} combine VLMs with globally rendered views for stronger spatial reasoning, while VLM-Grounder~\cite{VLM_Grounder}, SeqVLM~\cite{SeqVLM}, and SPAZER~\cite{SPAZER} further incorporate native RGB observations to preserve local semantic and spatial cues. Nevertheless, their generalization remains constrained by the training distribution of the underlying 3D segmentation backbone. UniGround addresses this gap by replacing segmentation-based perception with a training-free geometric perception pipeline, enabling zero-shot grounding across both language and scenes without 3D supervision.

% \begin{figure*}[!t]
%     \centering
%     \includegraphics[width=\textwidth]{figure/pipeline.png}
%     \caption{\textbf{Overview of UniGround.} Given raw scene observations and a language query, UniGround performs \textbf{Stage~1: Global Candidate Filtering} to retrieve a compact candidate set, followed by \textbf{Stage~2: Local Precision Grounding}, which localizes the target by jointly reasoning over global scene context for spatial verification and candidate-centric visual evidence for semantic identification.} \label{fig:pipeline}
% \end{figure*}

\vspace{-10pt}
\section{Methodology}
% In this section, we first define the task in Section~\ref{Sec: Problem_Formulation}. We then introduce UniGround, a dual-channel reasoning framework for zero-shot 3DVG that is applicable to any scene. As illustrated in Fig.~\ref{fig:pipeline}, the system takes raw scene observations as input, including first-person RGB-D sequences with camera poses, colored point clouds, and natural language instructions. The framework first performs global candidate filtering in Stage One~\ref{Sec: Stage_One} to isolate several potential objects from the entire scene that correspond to the instruction. Subsequently, in Stage Two~\ref{Sec: Stage_Two}, the system executes fine-grained localization on these candidates to precisely identify and bound the target object for the final grounded result.\

We formulate the task and introduce UniGround, a zero-shot 3DVG framework that addresses both bottlenecks (Fig.~\ref{fig:pipeline}). It first constructs topology-consistent candidates without dataset-trained 3D detectors, then grounds the referent by integrating global context, candidate-centric evidence, and closed-loop verification.

\vspace{-10pt}
\subsection{Problem Formulation}\label{Sec: Problem_Formulation}
The goal of the zero-shot 3D Visual Grounding (3DVG) task is to localize the target object within the 3D scene based on the textual description, without requiring scene-specific training or fine-tuning. The output is an oriented 3D bounding box (bbox) that specifies the target object's spatial location, dimensions, and orientation within the scene.
We represent the 3D scene as a colored point cloud, where each point has XYZ coordinates and RGB color attributes. This scene is accompanied by a sequence of first-person view images containing RGB, depth, and pose information. The system takes these visual inputs along with a natural language query to identify and bound the target object.

\vspace{-10pt}
\subsection{Stage One: Global Candidate Filtering}
\label{Sec: Stage_One}
\vspace{-12pt}
% This module is responsible for dividing the scene into distinct units and performing an initial selection of objects based on the instructions provided. Existing Zero-Shot 3D Visual Grounding approaches are contingent upon supervised 3D segmentation models to partition scenes and assign categorical labels. Although effective in controlled settings, these methods are inherently constrained by their reliance on supervised training data, which impedes their capacity to generalize to novel, unseen environments. To address this limitation, we draw inspiration from the Topological Clustering Strategy in HERO~\cite{wang2025hero}, which leverages the strengths of 2D models and spatial topology for 3D instance segmentation while assigning semantic categories for each instance through multi-angle RGB images and a design to fill in missing information.

\noindent
\begin{minipage}[t]{0.53\textwidth}
\vspace{0pt}
This module partitions the scene into distinct object instances and performs instruction-guided candidate filtering. Existing zero-shot 3DVG approaches typically rely on supervised 3D segmentation models for scene partitioning and category assignment, limiting their generalization to unseen environments. To overcome this dependency, we draw inspiration from the Topological Clustering Strategy in HERO~\cite{wang2025hero}, which combines 2D foundation models with spatial topology for 3D instance construction and derives instance semantics from multi-view RGB observations with missing-information completion.
\end{minipage}\hfill
\begin{minipage}[t]{0.44\textwidth}
    \vspace{-27pt}
    \centering
    \captionsetup{font=footnotesize}
    \includegraphics[width=\linewidth]{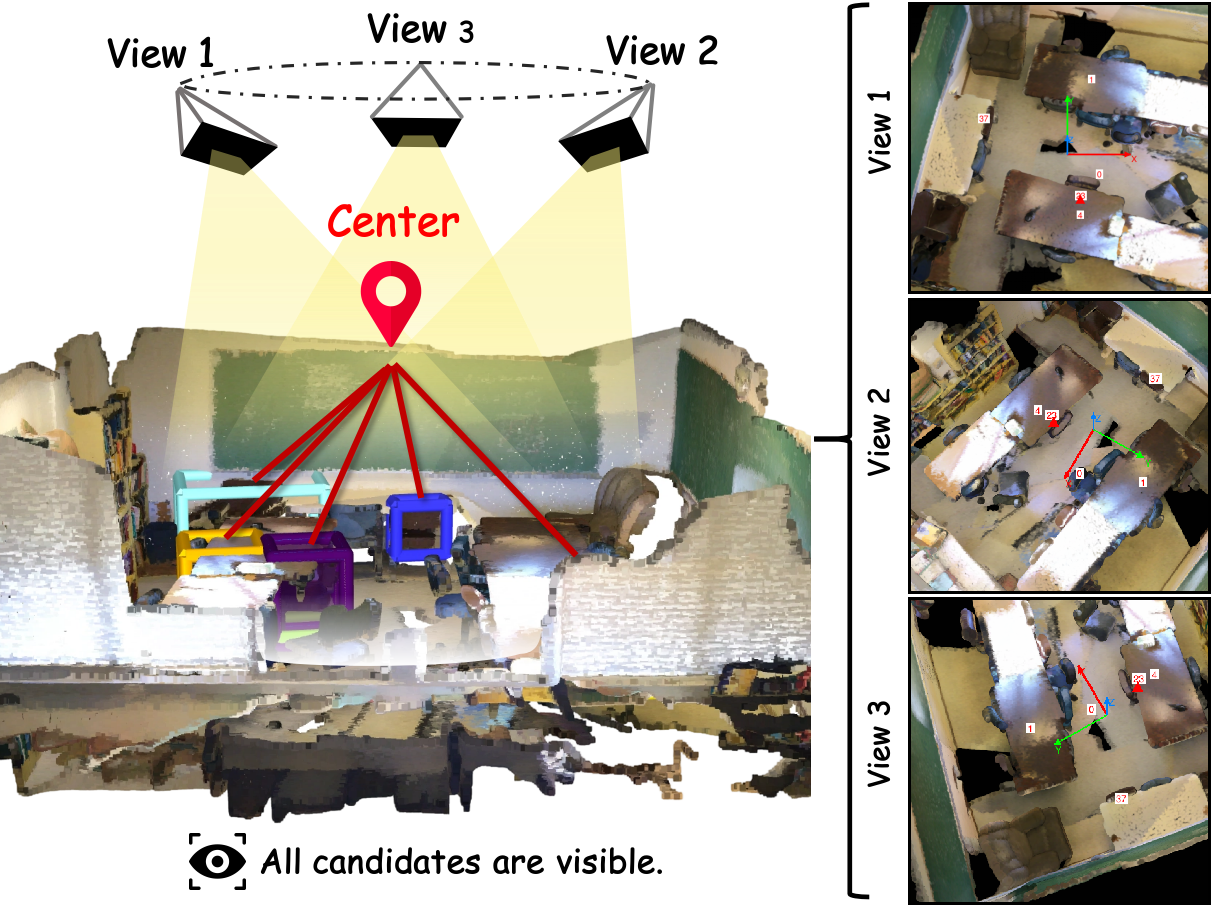}
    \captionof{figure}{Candidate-centered camera-orbit views for global spatial reasoning.}
    \label{fig:view}
\end{minipage}

\textbf{Segmentation.} We adopt a 2D-to-3D lifting strategy~\cite{wang2025hero,yin2024sai3d,takmaz2023openmask3d}: 2D instance masks are projected onto the 3D point cloud and aggregated into complete instances by merging geometrically connected regions with consistent semantics~\cite{papon2013voxel, zhang2023growsp,zhang2025logosp}. As shown in Fig.~\ref{fig:pipeline}, the input point cloud is first partitioned into superpoints via VCCS~\cite{papon2013voxel} and region growing~\cite{zhang2023growsp}, forming compact geometric primitives that preserve local continuity, planar smoothness, and appearance consistency. We then estimate the affinity between each pair of adjacent superpoints by identifying all camera views that jointly observe both regions, applying SAM~\cite{kirillov2023segment} to generate 2D instance masks, and evaluating joint visibility and semantic consistency. Formally,

\begin{equation}
C_{i,j} = \frac{1}{m} \sum_{k=1}^{m}
\underbrace{\frac{|i^{k}|_{\mathrm{vis}}\,|j^{k}|_{\mathrm{vis}}}{|i|\,|j|}}_{\text{joint visibility}}
\cdot
\underbrace{\frac{\mathbf{f}_i^k \cdot \mathbf{f}_j^k}
{\|\mathbf{f}_i^k\|_2\,\|\mathbf{f}_j^k\|_2}}_{\text{semantic consistency}},
\end{equation}
% where $i$ and $j$ are adjacent superpoints, $m$ is the number of views jointly observing both, $|i^k|_{\mathrm{vis}}$ and $|i|$ denote the visible and total pixel counts of superpoint $i$ in view $k$, and $\mathbf{f}_i^k \in \mathbb{R}^n$ is the feature vector of superpoint $i$ over the $n$ SAM-generated instance masks in view $k$, with $\mathbf{f}_j^k$ defined analogously. The resulting similarities drive a progressive merging scheme~\cite{yin2024sai3d}, where merging sensitivity is dynamically adjusted at each stage to fuse smaller regions first and progressively recover complete object instances.
where $(i,j)$ denotes adjacent superpoints, $m$ their co-visible views, $|i^k|_{\mathrm{vis}}$ and $|i|$ the visible and total pixel counts, and $\mathbf{f}_i^k\in\mathbb{R}^n$ the feature over $n$ SAM masks in view $k$ (analogously for $j$). These affinities guide adaptive progressive merging~\cite{yin2024sai3d}, which first fuses small regions and gradually recovers complete instances.

\begin{algorithm}[!t]
\caption{Contextual Semantic--Spatial Reasoning}
\label{alg:vlm_reasoning}
\small
\begin{algorithmic}[1]
\Require Query $q$; candidates $\mathcal{C}=\{1,\dots,K\}$; global renders $\mathcal{G}$; candidate views $\mathcal{V}=\{\mathcal{V}_i\}_{i=1}^K$
\Ensure Selected index $\hat{i}\in\mathcal{C}$ with explanation $e$

\State Define an axis--language mapping $\Phi$ consistent with the rendered coordinate frame

\Statex \textbf{Semantic reasoning (Naming-first)}
\For{$i=1$ to $K$}
    \State $\text{name}_i \leftarrow \textsc{VLM}(\mathcal{V}_i)$
\EndFor
\State $m \leftarrow \textsc{MatchTarget}(q,\{\text{name}_i\}_{i=1}^K)$

\Statex \textbf{Spatial reasoning}
\State $\mathcal{R} \leftarrow \textsc{ParseRelations}(q)$
\State $(\hat{i},e) \leftarrow \textsc{VLM}(\mathcal{G},\mathcal{R},\Phi,\mathcal{C})$

\Statex \textbf{Consistency-guided correction}
\If{$\hat{i}\notin\mathcal{C}$}
    \State $(\hat{i},e) \leftarrow \textsc{ConstrainedVLM}(\mathcal{G},\mathcal{R},\Phi,\mathcal{C})$
\ElsIf{\textsc{Inconsistent}$(m,\hat{i},q)$}
    \State $(\hat{i},e) \leftarrow \textsc{VLM}(\mathcal{G},\mathcal{V}_{\hat{i}},\mathcal{R},\Phi,\mathcal{C})$
\EndIf

\State $\hat{i} \leftarrow \textsc{ProjectToCandidates}(\hat{i},\mathcal{C})$
\State \Return $\hat{i},e$
\end{algorithmic}
\end{algorithm}

% \textbf{Semantic.} Geometric object proposals lack the semantic attributes required for query-driven selection. To obtain robust candidate embeddings, we introduce defect correction and multi-scale fusion. Since reconstruction artifacts and incomplete superpoint merging often produce ragged boundaries or missing surfaces, directly encoding the reconstructed geometry may introduce background clutter and occlusion noise. We therefore retrieve the corresponding RGB views using depth and camera poses, project the downsampled object point cloud onto image pixels, and use these projections as SAM prompts for conditional re-segmentation. The resulting masks provide cleaner object boundaries and are used to construct multi-scale crops that capture both local appearance and contextual cues. These crops are encoded by a Perception Encoder (PE)~\cite{bolya2025perception}, which maps visual and textual inputs into a shared embedding space. Multi-view features are averaged to suppress view-dependent noise, producing a stable semantic embedding for each candidate. At query time, the instruction is encoded by the same PE, and candidates with the highest cosine similarities are retained for downstream grounding.

\textbf{Semantic.} Geometric proposals lack the semantic attributes needed for query-driven selection. To obtain robust candidate embeddings, we apply defect correction and multi-scale fusion. Specifically, we project each downsampled object point cloud onto its corresponding RGB views using depth and camera poses, and use the projections as SAM prompts for re-segmentation. The refined masks reduce reconstruction noise and yield multi-scale crops containing both object appearance and surrounding context. A Perception Encoder (PE)~\cite{bolya2025perception} maps these crops and text into a shared embedding space, while multi-view averaging suppresses view-dependent noise. At inference, candidates with the highest cosine similarity to the PE-encoded instruction are retained for downstream grounding.

\subsection{Stage Two: Contextual Precision Grounding}
\label{Sec: Stage_Two}
\begin{table*}[t]
\centering
{%
  \scriptsize
  \setlength{\tabcolsep}{0.9mm}
  \renewcommand{\arraystretch}{1.14}
  \resizebox{\linewidth}{!}{%
  \begin{tabular}{lccccccccccc}
  \toprule
  \textbf{Method}
  & \multicolumn{2}{c}{\textbf{Agent}}
  & \multicolumn{2}{c}{\textbf{Prior}}
  & \multicolumn{2}{c}{\textbf{Unique}}
  & \multicolumn{2}{c}{\textbf{Multiple}}
  & \multicolumn{2}{c}{\textbf{ScanRefer}}
  & \textbf{ES} \\
  & \textbf{Rep.} & \textbf{VLM} & \textbf{Box} & \textbf{Cls.}
  & \textbf{.25} & \textbf{.50}
  & \textbf{.25} & \textbf{.50}
  & \textbf{.25} & \textbf{.50}
  & \textbf{.25} \\
  \midrule
  \rowcolor{unigroundgroup}
  \multicolumn{12}{c}{\textit{Fully Supervised Learning-Based Methods}} \\
  ScanRefer~\cite{chen2020scanrefer}
  & \ding{55} & \ding{55} & -- & --
  & 67.6 & 46.2 & 32.1 & 21.3 & 39.0 & 26.1 & 12.9 \\
  InstanceRefer~\cite{yuan2021instancerefer}
  & \ding{55} & \ding{55} & -- & --
  & 77.5 & 66.8 & 31.3 & 24.8 & 40.2 & 32.9 & -- \\
  3DVG-T~\cite{zhao20213dvg}
  & \ding{55} & \ding{55} & -- & --
  & 77.2 & 58.5 & 38.4 & 28.7 & 45.9 & 34.5 & -- \\
  BUTD-DETR~\cite{jain2022bottom}
  & \ding{55} & \ding{55} & -- & --
  & 84.2 & 66.3 & 46.6 & 35.1 & 52.2 & 39.8 & 22.1 \\
  Embodied Perceptron~\cite{wang2024embodiedscan}
  & \ding{55} & \ding{55} & -- & --
  & -- & -- & -- & -- & -- & -- & 25.7 \\
  \midrule
  \rowcolor{unigroundgroup}
  \multicolumn{12}{c}{\textit{Open-Vocabulary Zero-Shot Methods}} \\
  ZSVG3D~\cite{ZSVG3D}
  & \ding{55} & GPT-4V & \ding{51} & \ding{51}
  & 63.8 & 58.4 & 27.7 & 24.6 & 36.4 & 32.7 & -- \\
  VLM-Grounder~\cite{VLM_Grounder}
  & \ding{55} & GPT-4V & \ding{51} & \ding{51}
  & 66.0 & 29.8 & 48.3 & 33.5 & 51.6 & 32.8 & -- \\
  SeeGround~\cite{SeeGround}
  & \ding{55} & Qwen2-VL & \ding{51} & \ding{51}
  & 75.7 & 68.9 & 34.0 & 30.0 & 44.1 & 39.4 & 7.7 \\
  VoG~\cite{View_on_Graph}
  & \ding{55} & Qwen2-VL & \ding{51} & \ding{51}
  & 78.6 & 71.5 & 34.0 & 30.4 & 44.8 & 40.3 & -- \\
  SeqVLM~\cite{SeqVLM}
  & \ding{55} & Doubao-1.5 & \ding{51} & \ding{51}
  & 77.3 & 72.7 & 47.8 & 41.3 & 55.6 & 49.6 & 0.9 \\
  \midrule
  \rowcolor{unigroundgroup}
  \multicolumn{12}{c}{\textit{Open-World Zero-Shot Methods}} \\
  LERF~\cite{kerr2023lerf}
  & CLIP & \ding{55} & \ding{51} & \ding{51}
  & -- & -- & -- & -- & 4.8 & 0.9 & -- \\
  OpenScene~\cite{peng2023openscene}
  & CLIP & \ding{55} & \ding{51} & \ding{51}
  & 20.1 & 13.1 & 11.1 & 4.4 & 13.2 & 6.5 & -- \\
  LLM-G~\cite{dai2024llmg}
  & \ding{55} & GPT-3.5 & \ding{51} & \ding{51}
  & -- & -- & -- & -- & 14.3 & 4.7 & -- \\
  LLM-G~\cite{dai2024llmg}
  & \ding{55} & GPT-4T & \ding{51} & \ding{51}
  & -- & -- & -- & -- & 17.1 & 5.3 & -- \\
  \rowcolor{unigroundrow}
  \textbf{Ours}
  & PE & GPT-5 & \ding{55} & \ding{55}
  & 65.36 & 50.97 & 40.56 & 28.88
  & \best{46.1} & \best{34.1} & \best{28.7} \\
  \bottomrule
  \end{tabular}
  }
}
\caption{Evaluations of 3DVG on the ScanRefer validation set and a subset of the EmbodiedScan set within ARKitScenes. ``ES'' denotes EmbodiedScan; .25 and .50 denote Acc@0.25 and Acc@0.5, respectively.}
\label{tab:main}
\end{table*}

This stage addresses the evidence bottleneck by identifying the referent among Stage-One candidates. Global renderings preserve spatial relationships but lack fine-grained appearance cues, whereas candidate-centric RGB views capture discriminative details but provide limited context. Contextual Precision Grounding integrates both through Spatial Relationship Prompts and Candidate Visual Evidence, followed by closed-loop consistency checking to resolve conflicting predictions.

\textbf{Spatial Relationship Prompt.}
Point-cloud renderings are highly sensitive to camera pose, and suboptimal viewpoints may obscure scene structure or introduce misleading spatial cues. To recover global relations that are ambiguous in local crops, we render the scene from constrained multi-view camera poses centered on the candidate set (Fig.~\ref{fig:view}). Specifically, the camera orbits the mean candidate center according to
\begin{equation}
\mathrm{cam}(\theta)
= \bar{c} + [r\cos\theta,\; r\sin\theta,\; h]^{\top},
\quad
\bar{c}=\frac{1}{K}\sum_{i=1}^{K}c_i,
\end{equation}
where $K$ is the number of candidate boxes, $c_i$ is the center of the $i$-th candidate, and $\theta$, $r$, and $h$ denote the azimuth, orbit radius, and camera height, respectively. We enforce $r\ge r_{\min}$ and $h\ge h_{\min}$ to avoid degenerate viewpoints. Each rendering is further augmented with a global coordinate frame and unique candidate IDs to support directional reasoning and unambiguous object reference.

% \textbf{Candidate Visual Evidence.} 
% Directly rendering local object crops from point clouds into 2D images often yields unstable object appearances due to sparsity, occlusion, and viewpoint sensitivity, which can obscure discriminative visual attributes and lead to semantic loss. To mitigate this issue, we directly use the native RGB observations captured from the first-person view as the basis for constructing object-level local semantics. Specifically, for each candidate object, we select \( l \) views such that the object occupies a large pixel proportion in each image, while the pairwise spatial distances among the selected viewpoints are maximized, thereby preserving both visual detail and cross-view complementarity. Finally, to explicitly guide the model’s attention to the target object region, we overlay a bounding box on the object in each selected image, providing localized visual emphasis for more reliable semantic classification.

\textbf{Candidate Visual Evidence.}
Rendering local point-cloud crops into 2D images often produces unstable appearances due to sparsity, occlusion, and viewpoint sensitivity, obscuring discriminative attributes and causing semantic loss. We therefore construct object-level evidence directly from native first-person RGB observations. For each candidate, we select \(l\) views that maximize both its pixel coverage and the spatial diversity of camera viewpoints, preserving visual detail and cross-view complementarity. A bounding box is overlaid on the candidate in each view to focus the model on the relevant region and support reliable semantic classification.

\begin{figure*}[!t]
    \centering
    \includegraphics[width=\textwidth]{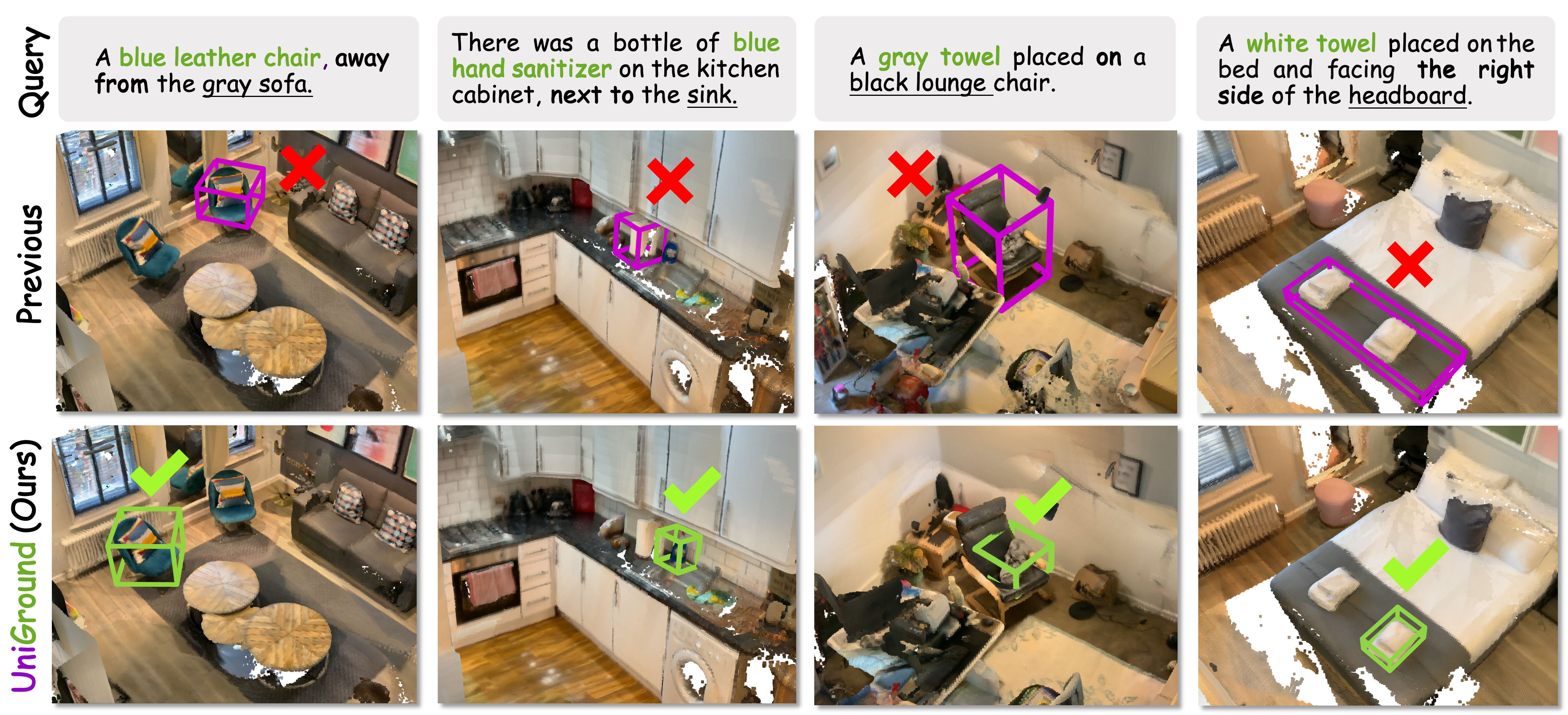}
    \caption{Qualitative comparison on ARKitScenes under domain shift. Purple/green boxes denote incorrect/correct predictions. UniGround improves candidate construction and disambiguation through topology-aware proposals and global--local evidence.}
    \label{fig:sim_vis}
\end{figure*}

\textbf{General VLM.} The VLM integrates complementary global and local evidence through closed-loop semantic--spatial reasoning. \emph{Semantic Identification} matches fine-grained descriptions from candidate-centric RGB views to the query, while \emph{Spatial Verification} evaluates inter-object relations and directional constraints in coordinate-aligned global renderings. \emph{Consistency Correction} then cross-checks both decisions and, when inconsistencies arise, jointly re-examines the global renderings, candidate views, and parsed relations before returning a valid candidate ID. Algorithm~\ref{alg:vlm_reasoning} formalizes this process.

\section{Experiments}
% In this section, we conduct extensive experiments to
% validate our proposed DreamNav in terms of its capability
% and feasibility. We first describe the experimental details
% (Sec. IV-A), then compare DreamNav with state-of-the-art
% methods in simulation environments (Sec. IV-B), and further
% evaluate its feasibility in real-world environments (Sec. IVC). Finally, we perform ablation studies on key components
% of DreamNav (Sec. IV-D).

\begin{figure*}[!t]
    \centering
    \includegraphics[width=\textwidth]{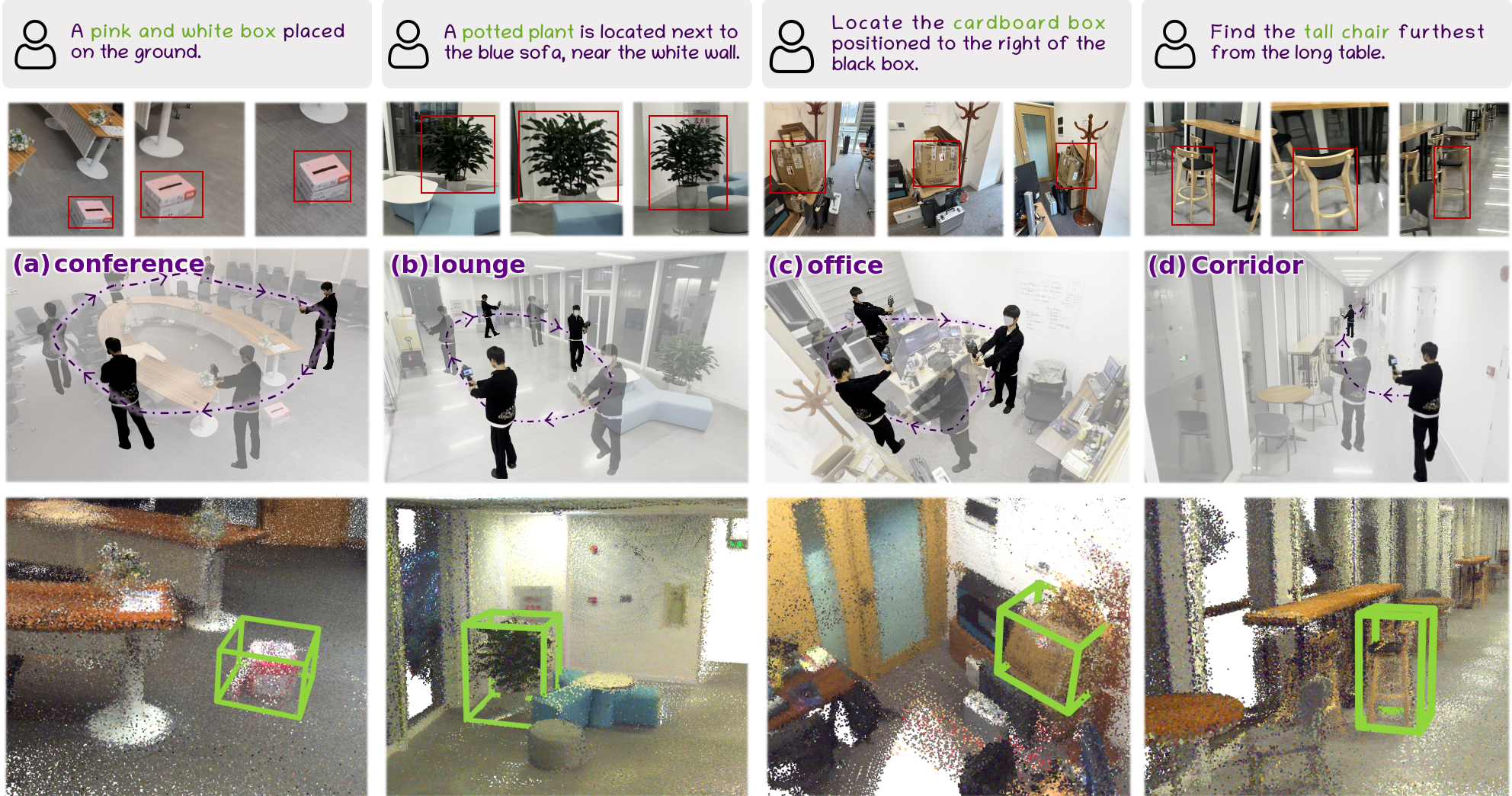}
    \caption{Qualitative results from the real-world indoor case study. UniGround localizes referred targets in previously unseen Conference, Lounge, Office, and Corridor environments under real-capture noise, imperfect reconstruction, and layout shifts.}
    \label{fig:real_world}
\end{figure*}

% Retain the final benchmark line on this page after the full-width float is placed.
\enlargethispage{\baselineskip}

\subsection{Experimental Details}

\begin{wrapfigure}{r}{0.44\columnwidth}
    \vspace{-6pt}
    \centering
    \includegraphics[width=\linewidth]{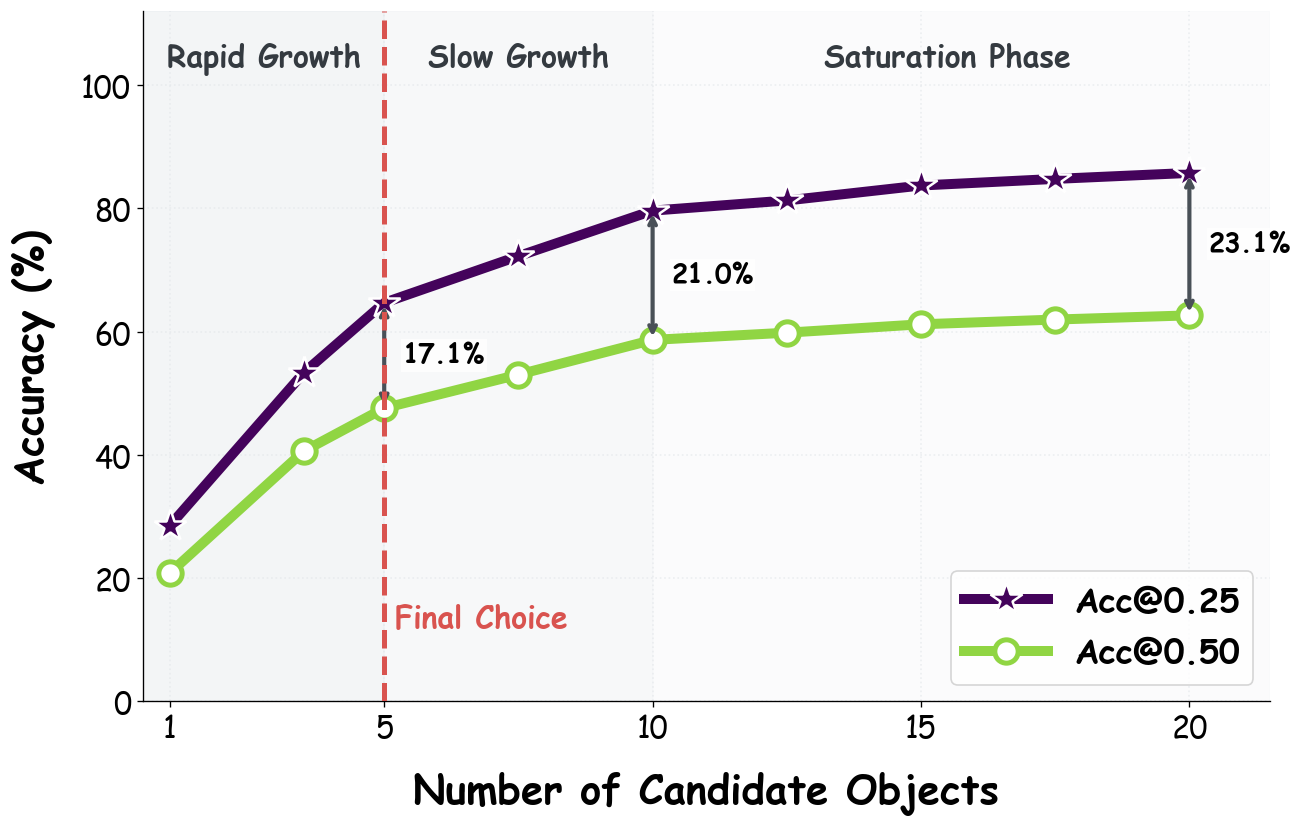}
    \captionsetup{font=footnotesize}
    \caption{Sensitivity to the number of candidates retained by Global Candidate Filtering.}
    \label{fig:ablation1}
    \vspace{-8pt}
\end{wrapfigure}

\textbf{Evaluation Benchmarks.} We evaluate on ScanRefer, a standard 3DVG benchmark containing 51,500 descriptions across 800 ScanNet scenes, including 9,508 validation queries. To assess cross-distribution generalization, we select over 6,000 tasks from 30 ARKitScenes in EmbodiedScan, which contains over 5,000 scans, 1 million language prompts, and 160,000 oriented 3D bounding boxes. We further construct a real-world benchmark across four indoor environments—Office, Lounge, Corridor, and Conference—with 25 queries per environment covering five target objects.

% \vspace{0.1cm}
\textbf{Implementation Details.}
In Stage One, instance segmentation adopts progressive-growing aggregation, with the similarity threshold linearly relaxed from 0.9 to 0.5 over five stages to consolidate regions at different scales. For semantic labeling, the multimodal Perception Encoder samples 10 viewpoints per object to generate semantic embeddings. We retain the top-$u$ candidates, with $u=5$ for ScanRefer and $u=10$ for EmbodiedScan. In Stage Two, Spatial Relationship Prompts use three global viewpoints to capture scene-level context, while Candidate Visual Evidence uses three local viewpoints per candidate for fine-grained semantic discrimination. Together, they enable robust localization through complementary global and local cues.

\subsection{Evaluation on Benchmark Environments}

Tab.~\ref{tab:main} compares UniGround with fully supervised, open-vocabulary zero-shot, and open-world zero-shot methods on ScanRefer and EmbodiedScan.

\textbf{UniGround achieves strong open-world grounding without dataset-specific priors.}
On ScanRefer, UniGround achieves 46.1/34.1 Acc@0.25/0.5, substantially outperforming the strongest open-world baseline, LLM-G with GPT-4T, by 29.0/28.8 percentage points. It also remains competitive with open-vocabulary methods that rely on predefined 3D bounding boxes and class priors. In particular, UniGround surpasses ZSVG3D, SeeGround, and VoG by 9.7, 2.0, and 1.3 percentage points in Acc@0.25, respectively. The results on the standard Unique/Multiple splits further demonstrate consistent grounding ability: UniGround obtains 65.36/50.97 Acc@0.25/0.5 on Unique instances and 40.56/28.88 on Multiple instances, despite using neither dataset-specific 3D proposals nor category priors.

\textbf{Cross-dataset generalization.}
On the evaluated ARKitScenes subset of EmbodiedScan, UniGround achieves 28.7\% Acc@0.25, outperforming SeeGround and SeqVLM by 21.0 and 27.8 percentage points, respectively, and Embodied Perceptron by 3.0 points. Unlike ScanNet-based evaluation, ARKitScenes introduces changes in sensor characteristics, scene layouts, reconstruction quality, and object distributions. The result therefore provides evidence of stronger robustness under the studied cross-dataset shift without task-specific 3D proposal supervision. Qualitative examples in Fig.~\ref{fig:sim_vis} further show that Global Candidate Filtering produces coherent candidates in cases where prior-dependent pipelines yield fragmented or incorrect proposals, while Contextual Precision Grounding uses complementary spatial and appearance evidence to resolve the retained candidates.

\subsection{Evaluation in Real-World Environments}

\begin{table*}[!t]
\centering
{%
\scriptsize
\setlength{\tabcolsep}{3.8mm}
\renewcommand{\arraystretch}{1.18}
\begin{tabularx}{\textwidth}{l*{4}{>{\centering\arraybackslash}X}}
\toprule
\textbf{Method} & \textbf{Conference} & \textbf{Lounge} & \textbf{Office} & \textbf{Corridor} \\
\midrule
SeeGround~\cite{SeeGround} & 2/25 & 0/25 & 5/25 & 0/25 \\
SeqVLM~\cite{SeqVLM} & 0/25 & 0/25 & 3/25 & 0/25 \\
\rowcolor{unigroundrow}
\textbf{Ours} & \best{10/25} & \best{13/25} & \best{15/25} & \best{5/25} \\
\bottomrule
\end{tabularx}
}
\caption{Evaluations of 3D visual grounding in four real-world environments.}
\label{tab:real_word}
\end{table*}

We evaluate UniGround against two representative open-source zero-shot 3DVG methods, SeeGround~\cite{SeeGround} and SeqVLM~\cite{SeqVLM}, in four real-world indoor environments: Conference, Lounge, Office, and Corridor. Each environment contains 25 language-query tasks involving five target objects.

\textbf{UniGround demonstrates robust real-world deployability.}
As reported in Tab.~\ref{tab:real_word}, UniGround consistently outperforms both baselines across four environments, improving the average success rate by 36 and 40 percentage points over SeeGround and SeqVLM, respectively. Even in the challenging Conference and Corridor environments, where both baselines nearly fail, UniGround maintains a 30.0\% average success rate. This robustness stems from its training-free two-stage design: Stage~1 constructs geometry-consistent candidates through superpoint aggregation and multi-view semantic verification, mitigating noise and reconstruction artifacts, while Stage~2 combines global spatial context with object-centric visual evidence to prevent error propagation. As shown in Fig.~\ref{fig:real_world}, UniGround reliably localizes targets even in cluttered scenes with imperfect reconstructions.

\subsection{Ablation Study}
To evaluate the contribution of each component within UniGround, we conduct ablation studies on 1,059 tasks randomly sampled from the ScanRefer dataset, covering 20 distinct ScanNet scenes.

\noindent\begin{minipage}{\linewidth}
  \centering
  {%
  \scriptsize
  \setlength{\tabcolsep}{2.6mm}
  \renewcommand{\arraystretch}{1.14}

  \begin{tabularx}{\linewidth}{>{\raggedright\arraybackslash}Xcc>{\raggedright\arraybackslash}Xcc}
    \toprule
    \rowcolor{unigroundgroup}
    \multicolumn{3}{c}{\textit{Backbone (without CoT)}} &
    \multicolumn{3}{c}{\textit{Prompts (GPT-5)}} \\
    \cmidrule(lr){1-3}\cmidrule(lr){4-6}
    \textbf{Model} & \textbf{.25} & \textbf{.5} &
    \textbf{Configuration} & \textbf{.25} & \textbf{.5} \\
    \midrule
    GPT-4o   & 41.6 & 31.6 & Spa. + Sem. & 54.0 & 40.3 \\
    GLM-4.5  & 42.8 & 31.2 & Sem. + Vis. & 51.6 & 38.1 \\
    Qwen-2.5 & 43.8 & 33.2 & Spa. + Vis. & 33.3 & 19.6 \\
    \rowcolor{unigroundrow}
    GPT-5 & \best{48.9} & \best{36.6} &
    \textbf{Spa. + Sem. + Vis.} & \best{56.2} & \best{41.0} \\
    \bottomrule
  \end{tabularx}
  }
  \captionof{table}{VLM and Stage 2 prompt ablations. Backbone: no CoT; prompts: GPT-5.}
  \label{tab:stage2_joint_ablation}
\end{minipage}

\textbf{Effect of VLM Backbones and Prompting Strategies.}
The VLM is the reasoning core of Stage~2, integrating spatial, semantic, and visual evidence for target grounding. As shown in Tab.~\ref{tab:stage2_joint_ablation}, with all CoT modules disabled, GPT-5 achieves 48.9\%/36.6\% Acc@0.25/0.5, outperforming Qwen-2.5 by 5.1/3.4 percentage points. Using GPT-5 as the backbone, combining Spatial, Semantic, and Visual CoT yields the best performance of 56.2\%/41.0\%, improving the baseline by 7.3/4.4 points. Removing Visual or Spatial CoT reduces performance to 54.0\%/40.3\% and 51.6\%/38.1\%, respectively, whereas removing Semantic CoT causes the largest drop to 33.3\%/19.6\%. These results highlight the importance of candidate-centric semantic evidence and the complementary contributions of all three CoT modules.

\textbf{Effect of the Number of Candidate Objects in Global Candidate Filtering.}
As the initial stage of our hierarchical framework, Global Candidate Filtering is critical because Stage~2 is restricted to the retained candidates. A larger candidate set improves global recall but also increases token consumption and inference latency. As shown in Fig.~\ref{fig:ablation1}, performance evolves in three phases as $N$ increases. In the \textit{Rapid Growth} phase ($N \leq 5$), Acc@0.25 rises sharply from 28.5\% to 64.7\%, demonstrating the importance of sufficient candidate coverage. The improvement then slows for $N \in [5,10]$ and saturates beyond $N=10$, with only a $\sim$6\% gain from $N=10$ to $20$. We therefore select $N=5$ as the best trade-off between localization accuracy and the computational overhead of Stage~2. These results suggest that the top-5 candidates effectively exclude most distractors, while the remaining errors mainly arise from inherent visual ambiguities.

% \textbf{Effect of Different VLM and Prompting Strategies.}
% As the reasoning core of Stage~2, the choice of VLM and the design of prompting strategies are critical for synthesizing multi-modal evidence.
% As summarized in Tab.~\ref{tab:combined_ablation}, we first examine the contribution of each prompting component under a fixed VLM setting.

% As shown in Table 3 (left), ablation experiments highlight the synergistic effect of spatial, semantic, and visual cues. Removing spatial relationship prompts leads to a performance drop from 54.0\% / 40.3\% to 51.6\% / 38.1\%, validating the necessity of explicit geometric frames for spatial reasoning. More critically, excluding candidate-centric semantic evidence results in a sharp decline to 33.3\% / 19.6\%, proving that global visual renderings alone are insufficient for fine-grained discrimination. Optimal performance of 56.2\% / 41.0\% is achieved only when all components are integrated within our structured reasoning protocol, underscoring their collective importance for precise grounding.

\section{Conclusion}
In this paper, we presented UniGround, a zero-shot 3D visual grounding framework designed around two sequential bottlenecks. Global Candidate Filtering addresses the candidate bottleneck by constructing topology-consistent object candidates without a dataset-trained 3D detector or predefined 3D box and category priors. Contextual Precision Grounding addresses the evidence bottleneck by jointly reasoning over global spatial context and candidate-centric visual evidence with closed-loop consistency checking. On ScanRefer, UniGround remains competitive without dataset-specific 3D proposals or category priors. Its results on the ARKitScenes subset of EmbodiedScan provide evidence of cross-dataset generalization, while the real-world indoor case study demonstrates robustness under the evaluated reconstruction noise and practical domain shifts. These results support candidate construction and contextual evidence reasoning as a promising direction for zero-shot 3D grounding in previously unseen indoor environments.

%In this paper, we present UniGround, a novel dual-channel reasoning framework that decouples geometric perception from semantic reasoning to achieve universal zero-shot 3D visual grounding. By replacing domain-specific 3D detection with a training-free Global Candidate Filtering stage, UniGround constructs topology-aware scene representations without relying on any 3D supervision, thus unlocking true open-world generalization. Within this framework, our Local Precision Grounding stage introduces a structured reasoning protocol that synthesizes candidate-centric visual evidence with global spatial relationship prompts, enabling the VLM to perform precise, hallucination-free localization. Experimental results on ScanRefer and EmbodiedScan demonstrate that UniGround not only produces strong performance among zero-shot methods but also outperforms fully supervised baselines in complex, out-of-distribution scenarios. Furthermore, we provide the first demonstration of a zero-shot 3DVG system successfully deployed in real-world environments, validating its robustness against domain shifts. We hope our work inspires future research into training-free, reasoning-centric paradigms for more generalizable embodied perception.

\clearpage
\bibliography{reference}

@String(ICLR  = {Int. Conf. Learn. Represent.})

@String(AAAI  = {AAAI})

@String(ICLR  = {ICLR})

@inproceedings{SeeGround,
  title={Seeground: See and ground for zero-shot open-vocabulary 3d visual grounding},
  author={Li, Rong and Li, Shijie and Kong, Lingdong and Yang, Xulei and Liang, Junwei},
  booktitle={Proceedings of the Computer Vision and Pattern Recognition Conference},
  pages={3707--3717},
  year={2025}
}

@inproceedings{VLM_Grounder,
  title={VLM-Grounder: A VLM Agent for Zero-Shot 3D Visual Grounding},
  author={Xu, Runsen and Huang, Zhiwei and Wang, Tai and Chen, Yilun and Pang, Jiangmiao and Lin, Dahua},
  booktitle={Conference on Robot Learning},
  pages={3961--3985},
  year={2025},
  organization={PMLR}
}

@inproceedings{SeqVLM,
  title={SeqVLM: Proposal-Guided Multi-View Sequences Reasoning via VLM for Zero-Shot 3D Visual Grounding},
  author={Lin, Jiawen and Bian, Shiran and Zhu, Yihang and Tan, Wenbin and Zhang, Yachao and Xie, Yuan and Qu, Yanyun},
  booktitle={Proceedings of the 33rd ACM International Conference on Multimedia},
  pages={3094--3103},
  year={2025}
}

@article{View_on_Graph,
  title={View-on-Graph: Zero-shot 3D Visual Grounding via Vision-Language Reasoning on Scene Graphs},
  author={Liu, Yuanyuan and Mei, Haiyang and Zhan, Dongyang and Zhao, Jiayue and Zhou, Dongsheng and Dong, Bo and Yang, Xin},
  journal={arXiv preprint arXiv:2512.09215},
  year={2025}
}

@inproceedings{ZSVG3D,
  title={Visual programming for zero-shot open-vocabulary 3d visual grounding},
  author={Yuan, Zhihao and Ren, Jinke and Feng, Chun-Mei and Zhao, Hengshuang and Cui, Shuguang and Li, Zhen},
  booktitle={Proceedings of the IEEE/CVF Conference on Computer Vision and Pattern Recognition},
  pages={20623--20633},
  year={2024}
}

@article{SPAZER,
  title={SPAZER: Spatial-Semantic Progressive Reasoning Agent for Zero-shot 3D Visual Grounding},
  author={Jin, Zhao and Tu, Rong-Cheng and Liao, Jingyi and Sun, Wenhao and Luo, Xiao and Liu, Shunyu and Tao, Dacheng},
  journal={arXiv preprint arXiv:2506.21924},
  year={2025}
}

@article{qi2025gpt4scene,
  title={Gpt4scene: Understand 3d scenes from videos with vision-language models},
  author={Qi, Zhangyang and Zhang, Zhixiong and Fang, Ye and Wang, Jiaqi and Zhao, Hengshuang},
  journal={arXiv preprint arXiv:2501.01428},
  year={2025}
}

@inproceedings{chen2020scanrefer,
  title={Scanrefer: 3d object localization in rgb-d scans using natural language},
  author={Chen, Dave Zhenyu and Chang, Angel X and Nie{\ss}ner, Matthias},
  booktitle={European conference on computer vision},
  pages={202--221},
  year={2020},
  organization={Springer}
}

@inproceedings{yuan2021instancerefer,
  title={Instancerefer: Cooperative holistic understanding for visual grounding on point clouds through instance multi-level contextual referring},
  author={Yuan, Zhihao and Yan, Xu and Liao, Yinghong and Zhang, Ruimao and Wang, Sheng and Li, Zhen and Cui, Shuguang},
  booktitle={Proceedings of the IEEE/CVF International Conference on Computer Vision},
  pages={1791--1800},
  year={2021}
}

@inproceedings{zhao20213dvg,
  title={3dvg-transformer: Relation modeling for visual grounding on point clouds},
  author={Zhao, Lichen and Cai, Daigang and Sheng, Lu and Xu, Dong},
  booktitle={Proceedings of the IEEE/CVF International Conference on Computer Vision},
  pages={2928--2937},
  year={2021}
}

@inproceedings{jain2022bottom,
  title={Bottom up top down detection transformers for language grounding in images and point clouds},
  author={Jain, Ayush and Gkanatsios, Nikolaos and Mediratta, Ishita and Fragkiadaki, Katerina},
  booktitle={European Conference on Computer Vision},
  pages={417--433},
  year={2022},
  organization={Springer}
}

@inproceedings{zhu20233dvista,
  title={3d-vista: Pre-trained transformer for 3d vision and text alignment},
  author={Zhu, Ziyu and Ma, Xiaojian and Chen, Yixin and Deng, Zhidong and Huang, Siyuan and Li, Qing},
  booktitle={Proceedings of the IEEE/CVF International Conference on Computer Vision},
  pages={2911--2921},
  year={2023}
}

@inproceedings{G3_LQ,
  title     = {{G$^3$-LQ}: Marrying Hyperbolic Alignment with Explicit Semantic-Geometric Modeling for 3D Visual Grounding},
  author={Wang, Yuan and Li, Yali and Wang, Shengjin},
  booktitle={Proceedings of the IEEE/CVF Conference on Computer Vision and Pattern Recognition},
  pages={13917--13926},
  year={2024}
}

@inproceedings{qian2024mcln,
  title={Multi-branch collaborative learning network for 3d visual grounding},
  author={Qian, Zhipeng and Ma, Yiwei and Lin, Zhekai and Ji, Jiayi and Zheng, Xiawu and Sun, Xiaoshuai and Ji, Rongrong},
  booktitle={European Conference on Computer Vision},
  pages={381--398},
  year={2024},
  organization={Springer}
}

@inproceedings{unal2024concretenet,
  title={Four ways to improve verbo-visual fusion for dense 3d visual grounding},
  author={Unal, Ozan and Sakaridis, Christos and Saha, Suman and Van Gool, Luc},
  booktitle={European Conference on Computer Vision},
  pages={196--213},
  year={2024},
  organization={Springer}
}

@inproceedings{guo2025tsp3d,
  title={Text-guided sparse voxel pruning for efficient 3d visual grounding},
  author={Guo, Wenxuan and Xu, Xiuwei and Wang, Ziwei and Feng, Jianjiang and Zhou, Jie and Lu, Jiwen},
  booktitle={Proceedings of the Computer Vision and Pattern Recognition Conference},
  pages={3666--3675},
  year={2025}
}

@inproceedings{kerr2023lerf,
  title={Lerf: Language embedded radiance fields},
  author={Kerr, Justin and Kim, Chung Min and Goldberg, Ken and Kanazawa, Angjoo and Tancik, Matthew},
  booktitle={Proceedings of the IEEE/CVF international conference on computer vision},
  pages={19729--19739},
  year={2023}
}

@inproceedings{peng2023openscene,
  title={Openscene: 3d scene understanding with open vocabularies},
  author={Peng, Songyou and Genova, Kyle and Jiang, Chiyu and Tagliasacchi, Andrea and Pollefeys, Marc and Funkhouser, Thomas and others},
  booktitle={Proceedings of the IEEE/CVF conference on computer vision and pattern recognition},
  pages={815--824},
  year={2023}
}

@inproceedings{dai2024llmg,
  title={Optimal scene graph planning with large language model guidance},
  author={Dai, Zhirui and Asgharivaskasi, Arash and Duong, Thai and Lin, Shusen and Tzes, Maria-Elizabeth and Pappas, George and Atanasov, Nikolay},
  booktitle={2024 IEEE International Conference on Robotics and Automation (ICRA)},
  pages={14062--14069},
  year={2024},
  organization={IEEE}
}

@article{liu2021deep,
  title={Deep view synthesis via self-consistent generative network},
  author={Liu, Zhuoman and Jia, Wei and Yang, Ming and Luo, Peiyao and Guo, Yong and Tan, Mingkui},
  journal={IEEE Transactions on Multimedia},
  volume={24},
  pages={451--465},
  year={2021},
  publisher={IEEE}
}

@article{liu2023raydf,
  title={RayDF: neural ray-surface distance fields with multi-view consistency},
  author={Liu, Zhuoman and Yang, Bo and Luximon, Yan and Kumar, Ajay and Li, Jinxi},
  journal={arXiv preprint arXiv:2310.19629},
  year={2023}
}

@inproceedings{liu2025unleashing,
  title={Unleashing the potential of multi-modal foundation models and video diffusion for 4d dynamic physical scene simulation},
  author={Liu, Zhuoman and Ye, Weicai and Luximon, Yan and Wan, Pengfei and Zhang, Di},
  booktitle={Proceedings of the Computer Vision and Pattern Recognition Conference},
  pages={11016--11025},
  year={2025}
}

@article{zhang2024navid,
  title={Navid: Video-based vlm plans the next step for vision-and-language navigation},
  author={Zhang, Jiazhao and Wang, Kunyu and Xu, Rongtao and Zhou, Gengze and Hong, Yicong and Fang, Xiaomeng and Wu, Qi and Zhang, Zhizheng and Wang, He},
  journal={arXiv preprint arXiv:2402.15852},
  year={2024}
}

@article{wang2025dreamnav,
  title={Dreamnav: A trajectory-based imaginative framework for zero-shot vision-and-language navigation},
  author={Wang, Yunheng and Fang, Yuetong and Wang, Taowen and Feng, Yixiao and Tan, Yawen and Zhang, Shuning and Liu, Peiran and Ji, Yiding and Xu, Renjing},
  journal={arXiv preprint arXiv:2509.11197},
  year={2025}
}

@inproceedings{zhou2024navgpt,
  title={Navgpt: Explicit reasoning in vision-and-language navigation with large language models},
  author={Zhou, Gengze and Hong, Yicong and Wu, Qi},
  booktitle={Proceedings of the AAAI Conference on Artificial Intelligence},
  volume={38},
  pages={7641--7649},
  year={2024}
}

@inproceedings{kong2023rethinking,
  title={Rethinking range view representation for lidar segmentation},
  author={Kong, Lingdong and Liu, Youquan and Chen, Runnan and Ma, Yuexin and Zhu, Xinge and Li, Yikang and Hou, Yuenan and Qiao, Yu and Liu, Ziwei},
  booktitle={Proceedings of the IEEE/CVF International Conference on Computer Vision},
  pages={228--240},
  year={2023}
}

@inproceedings{kong2023robo3d,
  title={Robo3d: Towards robust and reliable 3d perception against corruptions},
  author={Kong, Lingdong and Liu, Youquan and Li, Xin and Chen, Runnan and Zhang, Wenwei and Ren, Jiawei and Pan, Liang and Chen, Kai and Liu, Ziwei},
  booktitle={Proceedings of the IEEE/CVF International Conference on Computer Vision},
  pages={19994--20006},
  year={2023}
}

@article{li2022coarse3d,
  title={Coarse3d: Class-prototypes for contrastive learning in weakly-supervised 3d point cloud segmentation},
  author={Li, Rong and Cao, Anh-Quan and de Charette, Raoul},
  journal={arXiv preprint arXiv:2210.01784},
  year={2022}
}

@inproceedings{wang2024embodiedscan,
  title={Embodiedscan: A holistic multi-modal 3d perception suite towards embodied ai},
  author={Wang, Tai and Mao, Xiaohan and Zhu, Chenming and Xu, Runsen and Lyu, Ruiyuan and Li, Peisen and Chen, Xiao and Zhang, Wenwei and Chen, Kai and Xue, Tianfan and others},
  booktitle={Proceedings of the IEEE/CVF Conference on Computer Vision and Pattern Recognition},
  pages={19757--19767},
  year={2024}
}

@inproceedings{achlioptas2020referit3d,
  title={Referit3d: Neural listeners for fine-grained 3d object identification in real-world scenes},
  author={Achlioptas, Panos and Abdelreheem, Ahmed and Xia, Fei and Elhoseiny, Mohamed and Guibas, Leonidas},
  booktitle={European conference on computer vision},
  pages={422--440},
  year={2020},
  organization={Springer}
}

@inproceedings{li2025cityanchor,
  title={CityAnchor: City-scale 3D Visual Grounding with Multi-modality LLMs.},
  author={Li, Jinpeng and Wang, Haiping and Chen, Jiabin and Liu, Yuan and Dou, Zhiyang and Ma, Yuexin and Yang, Sibei and Li, Yuan and Wang, Wenping and Dong, Zhen and others},
  booktitle={ICLR},
  year={2025}
}

@inproceedings{zhang2023multi3drefer,
  title={Multi3drefer: Grounding text description to multiple 3d objects},
  author={Zhang, Yiming and Gong, ZeMing and Chang, Angel X},
  booktitle={Proceedings of the IEEE/CVF International Conference on Computer Vision},
  pages={15225--15236},
  year={2023}
}

@inproceedings{huang2025viewsrd,
  title={Viewsrd: 3d visual grounding via structured multi-view decomposition},
  author={Huang, Ronggang and Yang, Haoxin and Cai, Yan and Xu, Xuemiao and Zhang, Huaidong and He, Shengfeng},
  booktitle={Proceedings of the IEEE/CVF International Conference on Computer Vision},
  pages={9726--9736},
  year={2025}
}

@article{huang20253d,
  title={3d-r1: Enhancing reasoning in 3d vlms for unified scene understanding},
  author={Huang, Ting and Zhang, Zeyu and Tang, Hao},
  journal={arXiv preprint arXiv:2507.23478},
  year={2025}
}

@inproceedings{lin2025groundflow,
  title={Groundflow: A plug-in module for temporal reasoning on 3d point cloud sequential grounding},
  author={Lin, Zijun and He, Shuting and Tan, Cheston and Wen, Bihan},
  booktitle={Proceedings of the IEEE/CVF International Conference on Computer Vision},
  pages={28774--28784},
  year={2025}
}

@article{wang2025hero,
  title={HERO: Hierarchical Traversable 3D Scene Graphs for Embodied Navigation Among Movable Obstacles},
  author={Wang, Yunheng and Feng, Yixiao and Fang, Yuetong and Zhang, Shuning and Jing, Tan and Li, Jian and Jiang, Xiangrui and Xu, Renjing},
  journal={arXiv preprint arXiv:2512.15047},
  year={2025}
}

@article{huang2025openground,
  title={OpenGround: Active Cognition-based Reasoning for Open-World 3D Visual Grounding},
  author={Huang, Wenyuan and Wang, Zhao and Wei, Zhou and Huang, Ting and Zhao, Fang and Yang, Jian and Zhang, Zhenyu},
  journal={arXiv preprint arXiv:2512.23020},
  year={2025}
}

@article{takmaz2023openmask3d,
  title={Openmask3d: Open-vocabulary 3d instance segmentation},
  author={Takmaz, Ay{\c{c}}a and Fedele, Elisabetta and Sumner, Robert W and Pollefeys, Marc and Tombari, Federico and Engelmann, Francis},
  journal={arXiv preprint arXiv:2306.13631},
  year={2023}
}

@inproceedings{zhang2025logosp,
  title={Logosp: Local-global grouping of superpoints for unsupervised semantic segmentation of 3d point clouds},
  author={Zhang, Zihui and Dai, Weisheng and Wen, Hongtao and Yang, Bo},
  booktitle={Proceedings of the IEEE/CVF Conference on Computer Vision and Pattern Recognition},
  pages={1374--1384},
  year={2025}
}

@inproceedings{papon2013voxel,
  title={Voxel cloud connectivity segmentation-supervoxels for point clouds},
  author={Papon, Jeremie and Abramov, Alexey and Schoeler, Markus and Worgotter, Florentin},
  booktitle={Proceedings of the IEEE conference on computer vision and pattern recognition},
  pages={2027--2034},
  year={2013}
}

@inproceedings{zhang2023growsp,
  title={Growsp: Unsupervised semantic segmentation of 3d point clouds},
  author={Zhang, Zihui and Yang, Bo and Wang, Bing and Li, Bo},
  booktitle={Proceedings of the IEEE/CVF Conference on computer vision and pattern recognition},
  pages={17619--17629},
  year={2023}
}

@inproceedings{kirillov2023segment,
  title={Segment anything},
  author={Kirillov, Alexander and Mintun, Eric and Ravi, Nikhila and Mao, Hanzi and Rolland, Chloe and Gustafson, Laura and Xiao, Tete and Whitehead, Spencer and Berg, Alexander C and Lo, Wan-Yen and others},
  booktitle={Proceedings of the IEEE/CVF international conference on computer vision},
  pages={4015--4026},
  year={2023}
}

@inproceedings{yin2024sai3d,
  title={Sai3d: Segment any instance in 3d scenes},
  author={Yin, Yingda and Liu, Yuzheng and Xiao, Yang and Cohen-Or, Daniel and Huang, Jingwei and Chen, Baoquan},
  booktitle={Proceedings of the IEEE/CVF Conference on Computer Vision and Pattern Recognition},
  pages={3292--3302},
  year={2024}
}

@article{bolya2025perception,
  title={Perception encoder: The best visual embeddings are not at the output of the network},
  author={Bolya, Daniel and Huang, Po-Yao and Sun, Peize and Cho, Jang Hyun and Madotto, Andrea and Wei, Chen and Ma, Tengyu and Zhi, Jiale and Rajasegaran, Jathushan and Rasheed, Hanoona and others},
  journal={arXiv preprint arXiv:2504.13181},
  year={2025}
}

@article{FAST_LIO2,
  title={FAST-LIO2: Fast Direct LiDAR-inertial Odometry},
  author={Wei Xu and Yixi Cai and Dongjiao He and Jiarong Lin and Fu Zhang},
  journal={arXiv preprint arXiv:2107.06829},
  year={2021}
}

\clearpage
\appendix
\section*{Appendix}
\addcontentsline{toc}{section}{Appendix}

\section{Implementation and Evaluation Details}
\label{sec:implementation_details}

This section complements the method description in the main paper with the implementation choices needed to reproduce the two-stage pipeline. All components are used in inference mode: no benchmark-specific 3D detector, category vocabulary, or task-specific parameter training is introduced.

\subsection{Training-Free Candidate Construction}

For a scene containing $N_f$ RGB frames, Stage~1 reads frames with stride $\max(1,\lfloor N_f/1000\rfloor)$. This is an adaptive sampling stride rather than a hard cap on the number of processed frames. Before geometric aggregation, point coordinates are reordered from $(x,y,z)$ to $(x,z,y)$ and the last axis is sign-flipped so that observations from the reconstruction and image streams share a consistent frame.

The point cloud is initially decomposed into compact geometric primitives using VCCS~\cite{papon2013voxel} with a $0.05$ m voxel resolution. Progressive aggregation then merges adjacent regions by combining topology and multi-view semantic evidence. The similarity threshold is relaxed linearly from $0.9$ to $0.5$ over five stages, allowing highly consistent fragments to merge first and increasingly complete objects to form later. The implementation uses a maximum neighboring distance of 2, an $\ell_2$ similarity metric, distance decay $0.5$, spatial threshold $0.15$, and merge threshold 50; depth maps are interpreted in millimeters. These fixed settings are shared across scenes rather than tuned to individual language queries.

Semantic evidence is obtained from generic 2D foundation models. Semantic-SAM and SAM ViT-H~\cite{kirillov2023segment} supply complementary masks, while the Perception Encoder (PE)~\cite{bolya2025perception} maps visual regions and language into the shared embedding space used for candidate retrieval, as described in the main paper. For each object hypothesis, the implementation uses 10 encoding views, 10 random SAM rounds with five selected prompt points, and five nested crop levels with an expansion ratio of $0.1$. A view is retained when its visibility ratio exceeds $0.2$, and three representative evidence views are exported for downstream reasoning. Table~\ref{tab:stage1_configuration} summarizes the principal settings.

\begin{table}[H]
\centering
{%
\footnotesize
\setlength{\tabcolsep}{1mm}
\begin{tabular}{>{\raggedright\arraybackslash}p{0.18\textwidth}>{\raggedright\arraybackslash}p{0.27\textwidth}>{\raggedright\arraybackslash}p{0.49\textwidth}}
\toprule
\textbf{Component} & \textbf{Setting} & \textbf{Role} \\
\midrule
Frame sampling & Stride $\max(1,\lfloor N_f/1000\rfloor)$ & Adapts image processing density to the available RGB sequence. \\
Geometric initialization & VCCS resolution $0.05$,m & Forms topology-preserving primitives before instance aggregation. \\
Progressive aggregation & Five stages; similarity $0.9\!\rightarrow\!0.5$ & Merges confident fragments early and recovers more complete instances progressively. \\
Multi-view encoding & 10 views; visibility threshold $0.2$ & Accumulates semantic evidence across observations of an object hypothesis. \\
Mask prompting & 10 rounds; 5 points; 5 crop levels; expansion $0.1$ & Improves coverage across object scales and partial observations. \\
Exported visual evidence & 3 diverse views per candidate & Supplies compact, object-centric evidence to Stage~2. \\
\bottomrule
\end{tabular}
}
\caption{Principal Stage~1 implementation settings. They are fixed for the evaluated scenes and do not require task-specific training.}
\label{tab:stage1_configuration}
\end{table}

\subsection{Instruction-Guided Candidate Filtering}

After aggregation, each class-agnostic object hypothesis has a multi-view semantic embedding. The language query is encoded by the same PE model, and cosine similarity ranks all object hypotheses. We retain the top five candidates on ScanRefer and the top ten on EmbodiedScan, following the operating points stated in the main paper. For benchmark evaluation, each candidate region is converted directly from its constituent points to the center-and-extent representation $(c_x,c_y,c_z,d_x,d_y,d_z)$. Thus, language only filters already constructed geometric hypotheses; it does not supply a ground-truth category or box prior.

Ground-truth boxes are consulted only after ranking for evaluation and result export. In particular, if the largest IoU between the annotated target and any retained candidate is below $0.25$, the implementation records a Stage~1 miss by setting the target candidate ID to \texttt{null}. This separation prevents target annotations from influencing candidate ranking and also makes the two pipeline bottlenecks measurable: candidate recall at IoU $0.25$ or $0.5$ diagnoses whether Stage~1 has exposed the referent, whereas final grounding accuracy additionally tests whether Stage~2 selects it correctly.

\subsection{Evaluation Semantics}

The released evaluator supports two complementary views of performance. \emph{Complete} evaluation counts a Stage~1 miss as an incorrect end-to-end prediction and therefore reflects the behavior of the full system. \emph{Filtered} evaluation excludes those misses and measures Stage~2 conditionally on the referent being present in the candidate set. The latter is useful for diagnosing semantic--spatial reasoning, but it should not be interpreted as end-to-end accuracy. Reporting and interpreting these modes separately clarifies whether an error originates from candidate construction or final candidate selection.

\section{VLM Reasoning and Scene Rendering}
\label{sec:vlm_rendering_details}

\subsection{Details of Textual Prompt Design}

To enable stable and interpretable reasoning in the VLM, UniGround adopts a structured textual prompt that explicitly organizes visual inputs, grounding context, decision rules, and response constraints. Rather than directly predicting the target from mixed visual information, the prompt separates candidate-level semantic identification from scene-level spatial verification, consistent with the two-stage reasoning strategy in the main paper.

\textbf{Prompt design.}
Table~\ref{tab:agent_definition_prompt} summarizes the prompt interface used in UniGround. The prompt first defines the VLM as an object-grounding assistant and explains the roles of the stitched global render and candidate-centric images. It then provides grounding context, including candidate IDs, target and anchor hints, and 3D object metadata. Based on this input, the prompt instructs the model to first identify each candidate from its candidate-centric image and then verify the final target using spatial relations and global scene context. A key design element is the \emph{naming-first} strategy, which requires the model to assign semantic names to all candidates before making the final grounding decision.

The implementation instantiates this strategy with the system message ``You are a helpful assistant designed to identify objects based on images and descriptions.'' Candidate naming uses only the corresponding stitched candidate image and requests a descriptive phrase of three to eight words whenever possible. The model then checks those names against the query and uses the global renders and numerical 3D coordinates for verification. If no candidate name directly matches the target, or several names remain plausible, the prompt directs the model to identify the anchor first and resolve the target through the stated spatial relation. This fallback is important for open-vocabulary descriptions whose queried noun is visually uncommon or only recognizable from context.

To make directional language unambiguous, the prompt also states the coordinate convention used after rendering: left/right correspond to $+X/-X$, front/back to $-Y/+Y$, and above/below to $+Z/-Z$. These rules convert potentially view-dependent expressions into a consistent 3D reference frame. The complete user prompt is assembled dynamically: descriptions of global or candidate images are included only when those inputs are available, which also enables controlled component ablations without changing unrelated instructions.

\textbf{Structured output format.}
To support deterministic parsing, the prompt enforces a fixed response schema. As summarized in Table~\ref{tab:structured_output_fields}, the model is required to return six predefined fields, including the predicted object ID, an explanation of the decision, candidate naming results, the selected anchor ID, and the spatial relation used in the final step. Such a structured response format enables reliable and consistent extraction of the grounding result, while retaining important intermediate outputs produced during reasoning. In this way, it not only supports robust downstream parsing, but also exposes interpretable reasoning evidence for further analysis.

The response parser first reads this schema but also accepts a bare numeric ID as a conservative fallback, preventing harmless formatting variation from being counted as a grounding failure. API calls use a 300-second timeout and at most five attempts with exponential backoff (base 2) and random jitter. We leave sampling temperature and maximum output length unspecified so that the selected provider's defaults apply. These safeguards concern service reliability and output extraction; they do not inject additional scene or target information.

\textbf{Example reasoning trace.}
Table~\ref{tab:vlm_reasoning_output} shows a real inference trace produced under this prompt design. The example illustrates how the VLM first assigns semantic names to the candidate objects and then performs anchor-based spatial verification to reach the final prediction. In this way, the trace clearly reflects the intended reasoning flow of the prompt, in which candidate-level identification is followed by spatial verification based on anchor relations. Such traces provide an interpretable view of the reasoning process in both successful and failed cases.

\textbf{Design motivation.}
Overall, the structured prompt serves as an operational interface that translates the UniGround pipeline into a constrained VLM reasoning procedure. Instead of leaving the model to process heterogeneous inputs in an unconstrained manner, the prompt explicitly organizes candidate-centric semantics, global spatial context, reasoning rules, and structured outputs into a unified decision process. This explicit formulation helps reduce ambiguity during multimodal reasoning, encourages the model to use semantic and spatial evidence in a more coordinated way, and improves the reliability of open-world 3D visual grounding.

%%%%%%%%%%%%%%%表1 VLM prompt定义
\begin{table*}[t]
\centering
{%
\footnotesize
\setlength{\tabcolsep}{1mm}
\begin{tabular}{>{\raggedright\arraybackslash}p{0.16\textwidth}>{\raggedright\arraybackslash}p{0.50\textwidth}>{\raggedright\arraybackslash}p{0.27\textwidth}}
\toprule
\textbf{Module} & \textbf{Prompt Content} & \textbf{Purpose} \\
\midrule
\textbf{System Role} & You are a helpful assistant designed to identify objects based on images and descriptions. & Defines the VLM as an object-grounding agent. \\
\textbf{Visual Inputs} & A stitched global render provides scene-level spatial context (IDs and boxes), while stitched candidate images grouped by object ID provide object-centric visual evidence. & Combines global layout cues and local semantic evidence for robust grounding. \\
\textbf{Grounding Context} & Candidate IDs, target name, anchor name, optional anchor bbox, and each candidate's 3D center and size are provided. & Defines the grounding search space and explicit spatial metadata. \\
\textbf{Decision Protocol} & First identify candidates from candidate views, then verify the final target using global spatial relations and 3D coordinates; if ambiguous, resolve via anchor-first spatial reasoning. & Separates semantic identification from spatial verification and improves ambiguity handling. \\
\textbf{Response Format} & Predicted ID, Explanation, Candidate Names, Target name matches?, Anchor ID, Relation used. & Ensures deterministic parsing and interpretable reasoning traces. \\
\bottomrule
\end{tabular}
}
\caption{Agent definition prompt used in UniGround. The prompt explains visual inputs, grounding context, decision rules, and response constraints for the VLM.}
\label{tab:agent_definition_prompt}
\end{table*}

%%%%%%%%%%%%%%%%%%表3 VLM结构化输出
\begin{table}[t]
\centering
{%
\footnotesize
\setlength{\tabcolsep}{1mm}
\begin{tabular}{p{0.25\columnwidth}p{0.68\columnwidth}}
\toprule
\textbf{Field} & \textbf{Description} \\
\midrule
\texttt{Predicted ID} & The final object ID selected as the grounding result. \\
\texttt{Explanation} & A concise explanation describing the appearance and spatial cues used in the decision. \\
\texttt{Candidate Names} & Semantic names assigned to all candidate objects, formatted as \texttt{ID -> short phrase}. \\
\texttt{Target name matches?} & Indicates whether the selected candidate semantically matches the queried target category. \\
\texttt{Anchor ID} & The anchor object used during spatial reasoning, or \texttt{unknown} if no anchor is used. \\
\texttt{Relation used} & The spatial relation applied in the final decision (e.g., left of, near, behind). \\
\bottomrule
\end{tabular}
}
\caption{Structured output fields used for deterministic parsing of VLM reasoning results.}
\label{tab:structured_output_fields}
\end{table}

%%%%%%%%%%%%%%%表2：真实推理输出结果
\begin{table*}[t]
\centering
{%
\footnotesize
\setlength{\tabcolsep}{1mm}
\begin{tabular}{@{}p{0.96\textwidth}@{}}
\toprule
\textbf{Example VLM Output (Real Inference Trace)}\\
\midrule
\textbf{Query:} \texttt{a window that sits above the dresser cabinet. there is an air conditioner in the window.}\\[2pt]
\textbf{Ground-truth ID:} \texttt{22} \hspace{1em} \textbf{Predicted ID:} \texttt{22}\\[4pt]
\textbf{Predicted 3D Box (center \texttt{+} size):} \texttt{[0.41773347556591034, 2.9167184829711914, 1.1108443140983582, 0.4044213593006134, 1.1485633850097656, 0.8206886053085327]}\\[4pt]
\textbf{Ground-truth 3D Box (center \texttt{+} size):} \texttt{[0.39585648477077484, 3.0089060068130493, 1.1411845982074738, 0.3606673777103424, 0.9361202716827393, 0.7455374598503113]}\\[4pt]
\textbf{unique:} \texttt{null}\\[4pt]
\textbf{Explanation:} ID 22 visually shows a window with a window air conditioner installed. In the 3D data it is positioned higher (Z 1.11) and directly above the dresser/radiator cabinet (ID 20, Z 0.49), matching ``a window that sits above the dresser cabinet with an air conditioner in it.''\\[4pt]
\textbf{Candidate Names:} \texttt{7 -> beige wall section, 20 -> white radiator/dresser cabinet under window, 22 -> window with air conditioner, 41 -> boxy air conditioner unit on floor, 48 -> close-up of air conditioner unit on floor}\\[4pt]
\textbf{Target name matches?:} \texttt{true}\\
\textbf{Anchor ID:} \texttt{20}\\
\textbf{Relation used:} \texttt{above the dresser cabinet}\\
\bottomrule
\end{tabular}
}
\caption{Example VLM reasoning output in our structured format. This case correctly grounds the query to ID 22.}
\label{tab:vlm_reasoning_output}
\end{table*}

\subsection{Candidate-Centric Local Visual Evidence}
In Stage Two of UniGround, semantic identification is grounded in candidate-centric RGB observations rather than the global scene render. For each candidate, source views are first ranked by the number of pixels covered by its projected 3D region. Camera-center farthest-point sampling is then applied among the highly visible views: the highest-coverage view initializes the set, and subsequent views favor distinct viewing positions. This explicitly balances visibility and viewpoint diversity instead of selecting three nearly duplicate frames.

In practice, three complementary views are used to reduce viewpoint bias and capture distinctive appearance cues such as shape, texture, and nearby context. For each selected observation, the implementation saves both a tight object crop and its corresponding global RGB frame with a red bounding box; up to three such pairs form the candidate collage. This design helps the VLM focus on the intended instance while retaining enough local context to support semantic disambiguation. During inference, each stitched candidate image is associated with a candidate ID and serves as the primary visual evidence for candidate naming. As specified in Table~\ref{tab:structured_output_fields}, the VLM is required to assign semantic names to all candidates from these images before using the global render and 3D metadata for scene-level spatial verification.

%%%%%渲染的全局场景图相机视角选取原理%%%%%%
\subsection{Global Scene Rendering for Spatial Reasoning}

To support scene-level spatial verification, UniGround constructs candidate-set-centered global renderings that preserve the relative layout among retained objects. These renderings provide the geometric context used by the VLM to resolve spatial relations after candidate-level semantic identification, matching the camera-orbit formulation in the main paper.

\textbf{Candidate-Set-Centered Global Rendering.}
Given a set of $K$ retained candidate objects $\mathcal{O}$ with 3D bounding-box centers $\{\mathbf{c}_i\}_{i=1}^{K}$, we compute their common reference point as
\begin{equation}
\bar{\mathbf{c}} = \frac{1}{K}\sum_{i=1}^{K} \mathbf{c}_i,
\end{equation}
which serves as the camera look-at target. Parsed anchors remain visible in the render and are supplied to the VLM as relation evidence, but they do not replace the retained candidate set when defining the camera center. This distinction keeps all competing candidates within a common scene-level frame even when an anchor lies near the boundary of the scene or when a description contains more than one relational cue.

To obtain a global view that exposes the spatial layout of the scene, we adopt a tilted top-down viewing strategy. Let $\mathbf{d}_0=(0,0,1)$ denote the world vertical axis. The viewing direction is computed by applying tilt and azimuth rotations:
\begin{equation}
\mathbf{d} = R_{\text{azim}}(\phi)R_{\text{tilt}}(\theta)\mathbf{d}_0 .
\end{equation}

The camera distance is scaled according to the scene extent:
\begin{equation}
D = \alpha \cdot \max(\mathbf{b}_{\max}-\mathbf{b}_{\min}),
\end{equation}
where $\mathbf{b}_{\min}$ and $\mathbf{b}_{\max}$ denote the scene bounds. The camera center is computed as
\begin{equation}
\mathbf{e} = \bar{\mathbf{c}} + D\mathbf{d}.
\end{equation}

The horizontal and vertical components of $D\mathbf{d}$ correspond to the orbit radius and camera height in the main-paper formulation, with the same lower-bound safeguards against degenerate viewpoints. Varying the azimuth while fixing $\bar{\mathbf{c}}$ produces a family of coordinate-aligned views that exposes the spatial arrangement of the same candidate set without changing the grounding search space between views.

\textbf{Look-at View Transformation.}
After determining the camera center $\mathbf{e}$ and the look-at target $\bar{\mathbf{c}}$, we compute the camera extrinsic parameters using a standard look-at transformation. Let $\mathbf{up}=(0,0,1)$ denote the world up direction. The camera coordinate frame is defined as:
\begin{equation}
\begin{aligned}
\mathbf{z}_{axis}
&= \frac{\bar{\mathbf{c}}-\mathbf{e}}
        {\|\bar{\mathbf{c}}-\mathbf{e}\|},\\
\mathbf{x}_{axis}
&= \frac{\mathbf{up}\times\mathbf{z}_{axis}}
        {\|\mathbf{up}\times\mathbf{z}_{axis}\|},\\
\mathbf{y}_{axis}
&= \mathbf{z}_{axis}\times\mathbf{x}_{axis}.
\end{aligned}
\end{equation}

The rotation matrix is computed as:
\begin{equation}
R_c=
\begin{bmatrix}
\mathbf{x}_{axis}\\
\mathbf{y}_{axis}\\
\mathbf{z}_{axis}
\end{bmatrix}^{T},
\end{equation}
and the world-to-camera transformation is defined as:
\begin{equation}
\mathbf{x}_{cam}=R_c\mathbf{x}_{world}+T_c,
\quad
T_c=-R_c\mathbf{e}.
\end{equation}

\begin{figure*}[!t]
    \centering
    \includegraphics[width=\textwidth]{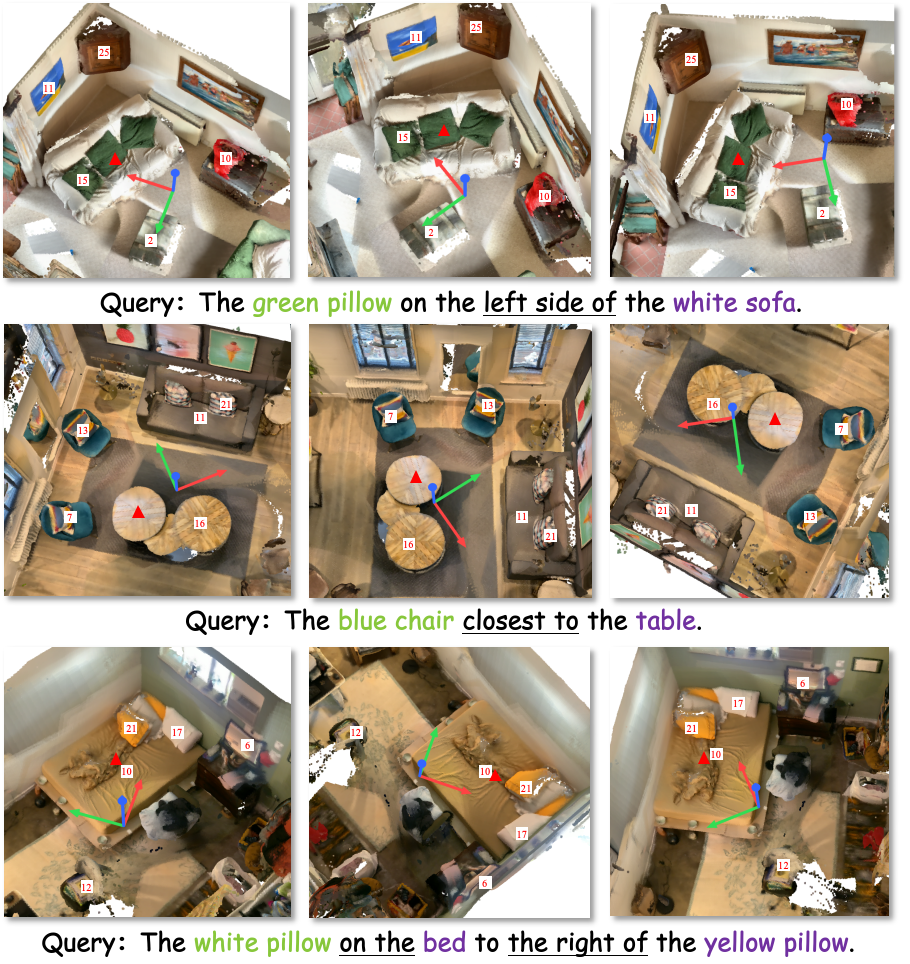}
    \caption{
    Examples of global multi-view scene renderings used as scene-level spatial context for the VLM. Each row shows a real grounding case with three candidate-set-centered rendered views from different azimuths. Candidate IDs are overlaid on the scene, the queried anchor is marked in red, and coordinate axes are visualized to support relation-aware reasoning.
    }
    \label{fig:app2}
\end{figure*}

Figure~\ref{fig:app2} shows examples of the resulting global multi-view renderings used during inference. For each query, three candidate-set-centered views are rendered to expose candidate layout, anchor position, and directional cues in a unified scene-level representation. Because the candidate IDs, 3D metadata, coordinate axes, and reference center remain fixed across azimuths, the three images are complementary observations of one geometric configuration rather than independent proposal sets.

Unless otherwise stated, the renderer outputs three $680\times680$ views at uniformly spaced azimuths of $0^\circ$, $120^\circ$, and $240^\circ$. We enable the top-down mode with a $15^\circ$ tilt, use the mean candidate center as the look-at target, and activate visibility-aware calibration so that all retained candidates remain in view. Before rendering, Stage~2 maps a box center $(x,y,z)$ to $(x,-z,y)$ and its extent $(d_x,d_y,d_z)$ to $(d_x,d_z,d_y)$. The mapping aligns reconstructed scenes with the coordinate directions stated in the prompt and therefore keeps the rendered axes, numerical metadata, and directional language consistent. Using three azimuths reduces the chance that an object--anchor relation hidden by overlap in one projection remains hidden in every view, while the shared coordinate convention prevents a change of viewpoint from changing the meaning of directional language.

\textbf{Visibility-Aware Camera Calibration.}
To improve object coverage in the rendered image, we optionally calibrate the camera intrinsics based on the projected extent of candidate object bounding boxes. 
Given a set of 3D points corresponding to candidate bounding box corners (or the scene point cloud), we project them into normalized device coordinates (NDC). 
Let $\mathbf{p}'$ denote the projection of a 3D point $\mathbf{p}$ in normalized device coordinates. 
The projected bounding region is estimated as
\begin{equation}
s=\frac{2}{\max(\mathbf{p}'_{\max}-\mathbf{p}'_{\min})}.
\end{equation}

The focal length and principal point are updated as:
\begin{equation}
f' = s f,\qquad
c' = s c - s\mathbf{m},
\end{equation}
where $\mathbf{m}$ is the center of the projected region. This calibration step improves object coverage and helps keep relevant candidates within the rendered view.

Calibration changes only the camera intrinsics used to display the already retained candidates. It does not alter their 3D coordinates, semantic scores, IDs, or ranking. The candidate set passed to Contextual Precision Grounding is therefore identical before and after rendering; calibration only makes the available geometric evidence more legible to the VLM. This preserves the separation stated in the main paper between Stage~1 candidate retrieval and Stage~2 evidence organization.

\textbf{Role in Spatial Reasoning.}
The global render serves as scene-level spatial context for the VLM. In conjunction with candidate-centric stitched images, it supports the second step of Stage Two, where the model verifies the final target using spatial relations and 3D layout cues. As illustrated in Fig.~\ref{fig:app2}, visual elements such as candidate IDs, anchor markers, and coordinate axes help the model interpret relations such as \emph{left of}, \emph{closest to}, \emph{in front of}, and \emph{above} within a consistent geometric frame. By presenting the candidates and anchors in a unified scene view, the global render allows the model to compare their relative positions more directly and reduces ambiguity that may arise from considering local object appearances alone. In this way, it provides the spatial evidence needed to disambiguate semantically similar candidates and supports more reliable relation-based verification.

\textbf{Design motivation.}
Overall, the global rendering strategy provides a geometry-aware scene representation that complements candidate-centric semantic evidence. By centering every view on the retained candidate set, preserving relative object layout, and improving candidate visibility, it supplies the VLM with reliable spatial context for relation verification. Anchors are interpreted inside this shared frame rather than used to redefine it. Rather than relying solely on candidate-level observations, the design explicitly introduces a scene-level reference frame in which object relations can be interpreted consistently across views. This implementation directly follows the main paper's decomposition: local RGB evidence supports semantic identification, while coordinate-aligned global evidence supports spatial verification and closed-loop consistency checking.

\FloatBarrier

\section{Additional Analyses}
\label{sec:additional_analyses}

\subsection{Protocol-Controlled Comparison and Transfer}

The main-paper comparison covers methods whose reported numbers may be obtained on different subsets or under different proposal mechanisms. To reduce one source of protocol variation, Table~\ref{tab:matched_subset} evaluates UniGround on the same 250-task subset used by VLM-Grounder~\cite{VLM_Grounder}. Under this task-matched protocol, UniGround reaches 66.4 Acc@0.25 and 48.1 Acc@0.5, improving over the reported VLM-Grounder results by 14.8 and 15.3 percentage points, respectively. This comparison controls the evaluated task subset, although the two systems retain their original backbones and pipelines; it therefore complements rather than replaces the broader comparison in the main paper.

The full ScanRefer validation results in the main paper remain the primary in-domain benchmark. The purpose of this smaller comparison is narrower: it verifies that the advantage is not created solely by evaluating UniGround and VLM-Grounder on different task lists. Because their proposal mechanisms, visual inputs, and VLM backbones are still different, the matched-subset result should be interpreted at the system level rather than as an isolated estimate of the contribution of Stage~1 or Stage~2. This distinction keeps the supplementary comparison consistent with the scope of the main table while providing a more controlled reference point.

\begin{table}[t]
\centering
{%
\footnotesize
\setlength{\tabcolsep}{2.5mm}
\begin{tabular*}{0.96\columnwidth}{@{\extracolsep{\fill}}lccc@{}}
\toprule
\textbf{Method} & \textbf{Tasks} & \textbf{Acc@0.25} & \textbf{Acc@0.5} \\
\midrule
VLM-Grounder & 250 & 51.6 & 32.8 \\
UniGround & 250 & \textbf{66.4} & \textbf{48.1} \\
\bottomrule
\end{tabular*}
}
\caption{Comparison on the same 250-task subset used by VLM-Grounder. Both accuracy columns are percentages.}
\label{tab:matched_subset}
\end{table}

\begin{table}[t]
\centering
{%
\footnotesize
\setlength{\tabcolsep}{2.5mm}
\begin{tabular*}{0.96\columnwidth}{@{\extracolsep{\fill}}lccc@{}}
\toprule
\textbf{Configuration} & \textbf{Global} & \textbf{Candidate} & \textbf{Structured} \\
\midrule
Full (Spa.+Sem.+Vis.) & Yes & Yes & Yes \\
No Global & No & Yes & Yes \\
No Candidate & Yes & No & Yes \\
Simple Prompt & Yes & Yes & No \\
\bottomrule
\end{tabular*}
}
\caption{Implementation mapping of the Stage~2 component ablations. Global denotes the scene render, Candidate denotes candidate-centric RGB, and Structured denotes the full reasoning prompt.}
\label{tab:ablation_mapping}
\end{table}

We further examine whether the pipeline transfers beyond ScanRefer-style descriptions. On 2,000 randomly sampled Nr3D tasks~\cite{achlioptas2020referit3d}, using the same box-IoU evaluation as in our other grounding experiments, UniGround obtains 48.2 Acc@0.25 and 36.9 Acc@0.5. Nr3D emphasizes human-written referring expressions and same-category distractors, so this result provides additional evidence that the training-free candidate and reasoning interface generalizes to a distinct language distribution without task-specific adaptation.

This diagnostic probes a different transfer axis from the ARKitScenes experiment in the main paper. ARKitScenes changes the scene source, sensor characteristics, reconstruction quality, and object distribution, whereas Nr3D changes the style and ambiguity of the referring expressions while remaining tied to ScanNet scenes. In both cases, the same two-stage principle is retained: query-independent geometric construction first exposes class-agnostic hypotheses, and language is used only to retrieve and reason over those hypotheses. The two evaluations are therefore complementary evidence rather than interchangeable claims of universal performance.

For the ARKitScenes subset of EmbodiedScan, the data converter maps the query target, referenced anchors, and their boxes into the same unified \texttt{task.json} interface used by the inference code. An optional ceiling filter removes points above the 95th height percentile, limiting the influence of overhead reconstruction artifacts on scene extent and rendering. Because the source annotations do not provide the ScanRefer \texttt{unique} attribute, converted ARKitScenes tasks conservatively default this field to false instead of inferring it from categories. These conversions standardize input representation without introducing target-dependent geometric filtering.

The conversion does not use the annotated target box to construct or rank candidates. As in the main-paper evaluation, the target annotation enters only when the predicted box is scored by IoU. UniGround retains ten candidates for EmbodiedScan, rather than the five-candidate ScanRefer operating point, and reports the end-to-end Acc@0.25 result over the evaluated subset. Thus, the reported 28.7\% measures the complete pipeline under the studied cross-dataset shift and does not condition on a successful Stage~1 match.

\subsection{Cross-Scene Analysis of Stage~1 Proposals}

\begin{figure*}[p]
    \centering
    \includegraphics[width=0.97\textwidth]{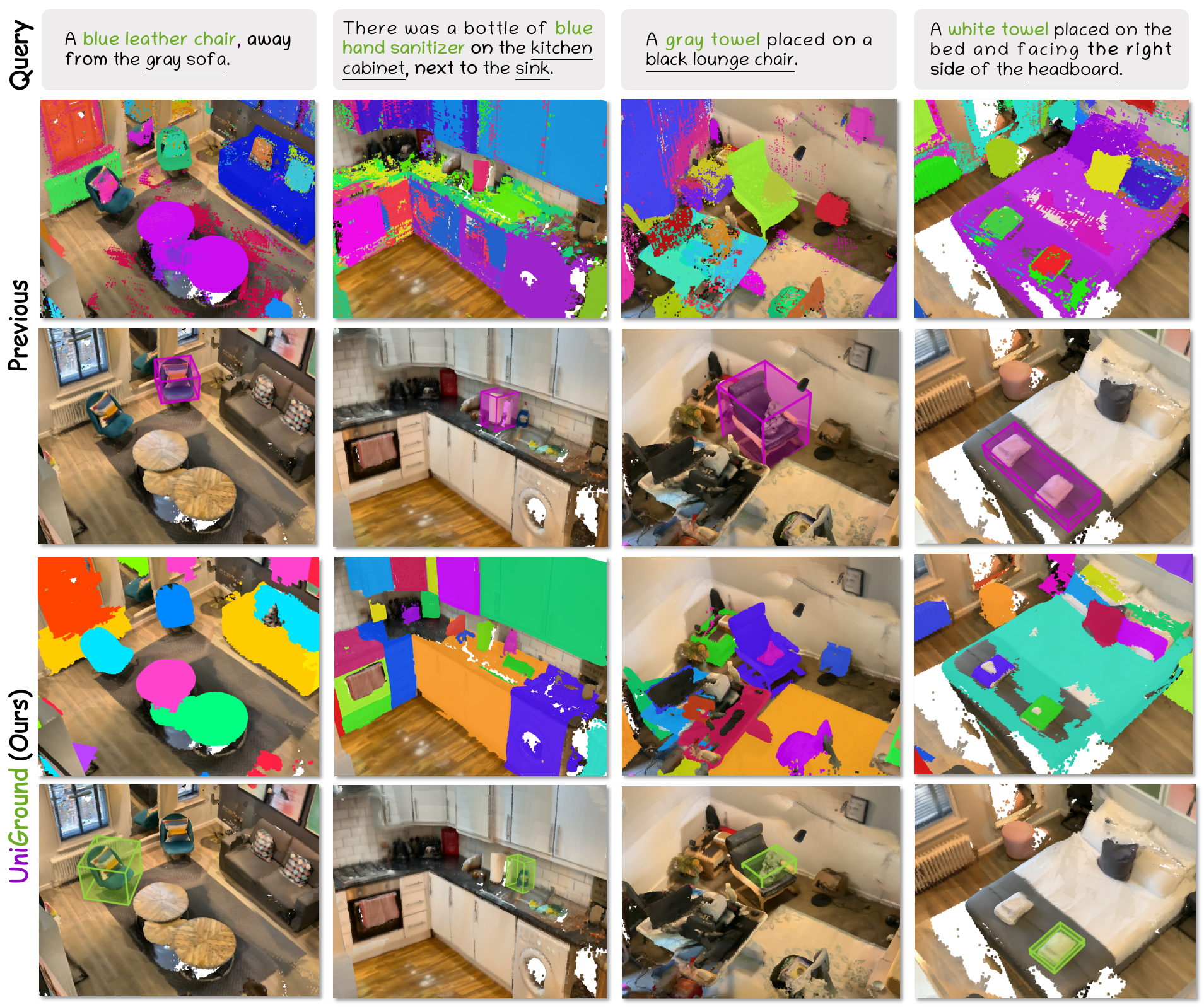}
    \caption{Qualitative comparison of Stage~1 candidate construction across four previously unseen indoor scenes. From top to bottom, each column shows the language instruction, the instance partition and selected proposal from a reference trained 3D-mask front end (purple), and the training-free UniGround partition and retained proposal (green). The examples highlight same-category confusion and small targets that are absorbed into larger supporting objects by the reference proposals, while UniGround exposes target-aligned candidates for downstream grounding.}
    \label{fig:stage1_cross_scene}
\end{figure*}

Figure~\ref{fig:stage1_cross_scene} provides a direct qualitative view of the candidate bottleneck under scene shift. The four columns span a lounge, a kitchen, a cluttered room, and a bedroom, none of which requires scene-specific training for UniGround. The queried referents also cover different geometric regimes: a same-category chair distractor, a small bottle on a crowded countertop, and two towels whose supporting furniture is substantially larger than the target. The upper pair of rows visualizes the instance partition and selected proposal from a reference trained 3D-mask front end, while the lower pair shows the corresponding training-free partition and retained proposal from UniGround.

The figure explicitly distinguishes proposal availability from final grounding. The colored partitions represent the object hypotheses generated by each Stage~1 front end, whereas the outlined regions indicate the proposal ultimately selected for the query in Stage~2. A correct grounding decision is possible only when Stage~1 exposes a sufficiently target-aligned candidate with accurate spatial coverage. The comparison therefore emphasizes both target visibility in the candidate set and proposal boundary quality, rather than attributing all failures solely to the final VLM-based selection.

\FloatBarrier

The comparison reveals a recurring difference in which objects become visible to the downstream grounding stage. In the lounge, the trained-mask proposal selects the wrong chair despite the explicit color cue, whereas UniGround preserves the blue leather chair specified by the instruction as a distinct candidate. On the kitchen counter, the reference partition is dominated by large cabinet and countertop regions, causing its selected proposal to cover a cluttered group of nearby objects. UniGround instead isolates the sanitizer bottle and exposes it as an individual candidate with a more precise boundary. The distinction is even more pronounced for the towels: the reference proposals follow the substantially larger supporting chair or bed-end region, while UniGround separates the small target surface from its support and produces a tight, target-aligned box. These examples illustrate a fundamental challenge for category- and dataset-conditioned 3D masks: the intended referent may be a small open-world object, a visually subordinate component, or an object whose geometry is easily absorbed into a larger and more familiar supporting structure.

Across the columns, the reference failures include both semantic selection errors and geometric under-segmentation. The lounge example contains a plausible same-category proposal but associates the instruction with the wrong instance; the bottle and towel examples are more restrictive because the target is absorbed into a larger region before language reasoning begins. UniGround's advantage in these cases is consequently not limited to assigning a better label to an unchanged box. Its class-agnostic construction makes a distinct, localized target hypothesis available, after which the PE-based shortlist can use the referring expression to retain it for Contextual Precision Grounding.

UniGround avoids committing to a fixed 3D category inventory during candidate construction. Instead, topology-preserving primitives, multi-view 2D mask agreement, and progressive aggregation determine instance boundaries before query-conditioned semantic ranking. Consequently, the Stage~2 reasoner receives a candidate corresponding to the described object rather than being forced to choose among proposals that have already absorbed or omitted it. The figure is intended as a qualitative proposal-level diagnostic rather than a controlled replacement of Stage~1 inside an otherwise fixed end-to-end system; nevertheless, the consistent target-aligned candidates across these four scene types illustrate the cross-scene robustness enabled by the training-free construction strategy.

This interpretation is consistent with, but does not replace, the quantitative evidence in the main paper. The ARKitScenes results evaluate the complete system under a defined cross-dataset protocol, whereas Figure~\ref{fig:stage1_cross_scene} makes the mechanism behind that robustness visually inspectable in additional scene types. Together they support the more specific conclusion that removing a dataset-trained 3D proposal front end reduces one important source of failure under scene shift; they do not imply that every remaining grounding error is solved once a target-aligned candidate is present.

\textbf{Mechanistic interpretation.}
The qualitative difference follows the ordering of operations in Global Candidate Filtering. VCCS and region growing first decompose the point cloud into compact primitives using local topology rather than a learned 3D category decision. Multi-view masks then provide agreement evidence only for regions that are geometrically adjacent and jointly visible, and progressive aggregation merges the most consistent fragments before relaxing the threshold to recover more complete instances. This ordering allows a small region to remain distinct when its topology and repeated 2D observations disagree with a much larger support object. In the bottle and towel columns, this property is visible in the finer boundary between the referent and the counter, chair, or bed. The result is consistent with the main-paper formulation of Stage~1 as topology-consistent, class-agnostic candidate construction; it should not be read as learning a new category model from these scenes.

Language enters only after these geometric hypotheses have been formed. PE embeds the multi-view evidence for each hypothesis and the instruction in a shared space, and cosine similarity retains the highest-ranked candidates. Consequently, the green boxes in Figure~\ref{fig:stage1_cross_scene} represent query-relevant selections from an already constructed partition, not boxes generated from target annotations or fitted directly to the query. This causal separation is important for the zero-shot setting: the same scene parsing can support different referring expressions, while language changes the shortlist rather than rewriting the underlying geometry. It also explains why a visually coherent but semantically irrelevant region can be discarded without imposing a predefined 3D label inventory on every proposal.

The diagnostic also clarifies the boundary between the two bottlenecks analyzed in the main paper. Stage~1 succeeds when a target-aligned hypothesis survives the shortlist; Stage~2 must still identify that hypothesis from candidate-centric appearance and verify its spatial relations in the global render. A coherent proposal therefore removes a hard candidate-visibility failure but does not guarantee the final answer. Same-class ambiguity can remain after the target is retained, as reflected by the lower performance on the Multiple split reported in the main paper. UniGround addresses these failure sources with different components: training-free scene parsing improves what the VLM is allowed to consider, while complementary global and local evidence improves how it selects among the retained candidates. Figure~\ref{fig:stage1_cross_scene} isolates the former contribution and thereby provides a visual counterpart to the end-to-end cross-dataset evaluation.

This qualitative analysis also motivates the use of a compact candidate set rather than a dense list of all geometric fragments. Once the desired object is represented by a localized proposal, passing many visually redundant neighboring regions to the VLM mainly increases context length and can introduce additional same-category distractors. The retained shortlist is therefore intended to preserve target availability while keeping the final reasoning problem small enough for detailed semantic naming and relation checking.

\FloatBarrier

\begin{table}[t]
\centering
{%
\footnotesize
\setlength{\tabcolsep}{1mm}
\begin{tabular}{ccc}
\toprule
\textbf{Candidates} & \textbf{Latency (s)} & \textbf{Tokens} \\
\midrule
3 & 30.61 & 1927.67 \\
5 & 38.95 & 5737.66 \\
7 & 49.09 & 7200.53 \\
10 & 59.82 & 9597.21 \\
\bottomrule
\end{tabular}
}
\caption{Measured Stage~2 inference cost as the number of retained candidates increases.}
\label{tab:candidate_efficiency}
\end{table}

\subsection{Evidence Organization and Controlled Ablations}

The Stage~2 ablations in the main paper correspond directly to the released inference switches, as detailed in Table~\ref{tab:ablation_mapping}. This mapping isolates three aspects of the reasoning interface: scene-level spatial evidence, candidate-centric RGB evidence, and the structured naming-and-verification protocol. No Global removes only the rendered scene-level context, No Candidate removes native RGB appearance evidence for candidate naming, and Simple Prompt retains both visual inputs but removes the structured reasoning constraints. In the simple-prompt condition, the system receives the query and candidate metadata and is asked only to identify an object ID and explain the choice; the explicit naming-first, anchor fallback, directional convention, and six-field output constraints are removed.

Concretely, its system message is ``You are a helpful assistant. Identify the object ID that matches the description.'' The user message contains only \texttt{Query: \{query\}}, \texttt{Candidate Objects: \{objects\_info\}}, and the request ``Please identify which object ID best matches the query and explain why.'' This compact form makes the ablation reproducible and distinguishes prompt simplification from removal of visual evidence.

\begin{figure*}[p]
    \centering
    \includegraphics[width=0.80\textwidth]{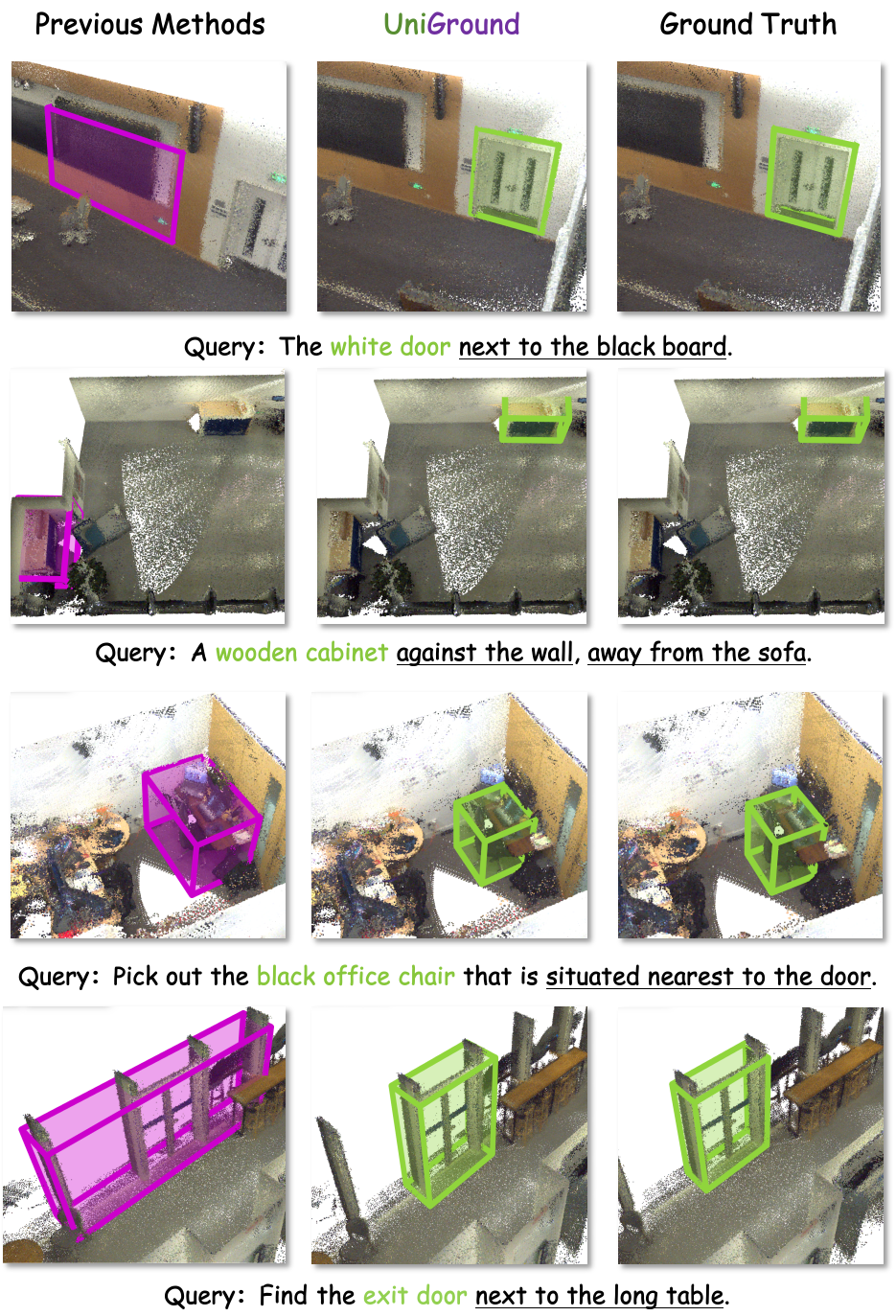}
    \caption{Representative real-world grounding outputs from comparison predictions, UniGround, and the annotated targets.}
    \label{fig:app3}
\end{figure*}
\FloatBarrier

As a more stringent diagnostic, we also evaluated a deliberately naive Stage~2 interface using the same five retained candidates, the original query, a plain prompt, and one raw scene rendering. It achieves 20.5/13.4 Acc@0.25/0.5. Since the candidate set is held fixed, the gap from the complete system cannot be attributed to better candidate recall. Instead, it shows that simply exposing a VLM to candidates and an unstructured scene image is insufficient: separating local identification from global verification and presenting coordinate-consistent evidence are consequential parts of the method.

\begin{table*}[t]
    \centering
    {%
    \footnotesize
    \setlength{\tabcolsep}{1mm}
    \begin{tabular}{>{\raggedright\arraybackslash}p{0.18\textwidth}>{\raggedright\arraybackslash}p{0.76\textwidth}}
        \toprule
        \textbf{Target Object}  & \multicolumn{1}{c}{\textbf{Instruction}}\\
        \midrule
        \multirow{5}{=}{cardboard box}  & Locate the \textbf{cardboard box} positioned to the right of the \underline{black box}. \\
        & Identify the \textbf{cardboard box} situated on the right side of the \underline{black box}. \\
        & Look for the \textbf{cardboard box} that is placed immediately to the right of the \underline{black box}. \\
        & Spot the \textbf{cardboard box} sitting adjacent to the right of the \underline{black box}. \\
        & Find the \textbf{cardboard box} located near the right-hand side of the \underline{black box}. \\
        \midrule
        \multirow{5}{=}{black office chair}  & Find the \textbf{black office chair} closest to the \underline{door}. \\
        & Pick out the \textbf{black office chair} that is situated nearest to the \underline{door}. \\
        & Locate the \textbf{black office chair} positioned right by the \underline{door}. \\
        & Find the \textbf{black office chair} that stands closest to the \underline{doorway}. \\
        & Search for the \textbf{black office chair} placed in the immediate vicinity of the \underline{door}. \\
        \midrule
        \multirow{5}{=}{gray blackout curtain}  & Find the \textbf{gray blackout curtain} next to the \underline{window}. \\
        & Identify the \textbf{gray blackout curtain} hanging beside the \underline{window}. \\
        & Locate the \textbf{gray blackout curtain} that is attached to the \underline{window frame}. \\
        & Find the \textbf{gray blackout curtain} positioned right next to the \underline{window}. \\
        & Look for the \textbf{gray blackout curtain} covering the area adjacent to the \underline{window}. \\
        \midrule
        \multirow{5}{=}{black chair}  & Find the \textbf{black chair} by the \underline{window}. \\
        & Search for the \textbf{black chair} positioned near the \underline{window}. \\
        & Locate the \textbf{black chair} that is sitting right by the \underline{window}. \\
        & Identify the \textbf{black chair} placed in front of or next to the \underline{window}. \\
        & Find the \textbf{black chair} situated in the corner by the \underline{window}. \\
        \midrule
        \multirow{5}{=}{silver box}  & Find the \textbf{silver box} next to a \underline{gray lounge chair}. \\
        & Identify the \textbf{silver box} sitting on the floor next to the \underline{gray lounge chair}. \\
        & Locate the \textbf{silver box} placed beside the \underline{gray lounge chair}. \\
        & Find the \textbf{silver box} that is positioned right adjacent to the \underline{gray lounge chair}. \\
        & Look for the \textbf{silver box} situated in the space next to the \underline{gray lounge chair}. \\
        \bottomrule
    \end{tabular}
    }
    \caption{Examples of task instructions for the Office scene in real-world environments. For each target, five diverse natural language queries are provided to evaluate the model's grounding consistency. Bold text denotes the target object, while underlined text indicates the spatial anchor used for relational reasoning.}
\label{tab:appendix_task}
\end{table*}

\subsection{Candidate Budget and Inference Cost}

The candidate count controls a practical trade-off. Increasing it can recover targets omitted by a narrow Stage~1 shortlist, but it also enlarges the multimodal context processed in Stage~2. Table~\ref{tab:candidate_efficiency} reports the measured latency and token usage from the candidate-budget diagnostic. Both costs grow monotonically with the number of retained objects. Relative to ten candidates, the five-candidate setting reduces latency by approximately 34.9\% and token usage by 40.2\%. Together with the accuracy trend in the main paper, this supports $N=5$ as a budget-aware default for ScanRefer, rather than claiming it is the accuracy-maximizing setting.

The cost increase is not purely textual. Each additional candidate may contribute candidate-centric crops, object metadata, and a larger set of pairwise spatial alternatives for the VLM to compare. This makes the candidate budget an interface parameter as well as a retrieval parameter: it determines how much visual evidence can be inspected in one reasoning call and how much ambiguity must be resolved by the structured prompt. The reported numbers therefore complement accuracy by quantifying the practical overhead of exposing a broader hypothesis set.

\section{Experimental Setup in Real-World Environments}

\subsection{Hardware Overview}
In our real-world experiments, we used the GeoScan S1, a compact handheld 3D laser scanner developed by TaoBotics in collaboration with Tongji University and Northwestern Polytechnical University for real-time scene reconstruction. It features multi-sensor fusion, including a Livox Mid-360 LiDAR with a 360° field of view, an Intel D435i depth camera, an ICM40609 IMU, and a T-RTK UM982 Mobile RTK GPS module, ensuring precise measurements with an accuracy of up to 5 cm. The system supports real-time SLAM for dynamic 3D point cloud generation and can cover large areas of up to 200,000 square meters. With a processing speed of 200,000 points per second, the GeoScan S1 offers reliable data output in formats like PCD, LAS, and PLY, which can be directly integrated into robotic navigation and environmental modeling applications. It operates under Ubuntu 20.04 and ROS, providing a flexible platform for research experiments. The device is powered by a 88.8Wh battery, offering 3 to 4 hours of operational time and includes wireless connectivity options like WiFi and Bluetooth for seamless data transfer.

\subsection{Experimental Pipeline Overview}
The experimental workflow begins with real-time spatial reconstruction using the GeoScan S1, employing a customized version of the Fast-LIO2~\cite{FAST_LIO2} algorithm. This process enables the synchronized acquisition of dense 3D point clouds, RGB-D data, and high-precision trajectory poses. To ensure data quality for downstream tasks, the raw point clouds are preprocessed using CloudCompare, where manual filtering and cropping are performed to eliminate noise and outliers. Subsequently, the refined data is processed through the proposed UniGround framework. In Stage 1, the system performs large-scale scene parsing to generate instance-level segmentations and their corresponding semantic embeddings~\cite{bolya2025perception, kirillov2023segment}. Based on predefined natural language queries, a set of candidate objects is filtered from the global scene. Finally, in Stage 2, these candidates are jointly reasoned with the target queries to achieve cross-modal alignment, ultimately outputting the unique target object that matches the specific task requirements. Figure~\ref{fig:app3} presents representative grounding outputs in multiple real-world scenes. The successful UniGround predictions illustrate how the combination of candidate-centric appearance and global spatial context can recover the intended referent despite viewpoint changes, clutter, and linguistically similar distractors. The examples are qualitative rather than a standalone comparison protocol; quantitative evaluation of the four environments is reported in the main paper.

\subsection{Task and Instruction Configuration}

To evaluate the robustness and spatial reasoning capabilities of our method in real-world environments, we selected four representative indoor scenes: Conference Room, Lounge, Office, and Corridor. Within each scene, five distinct target objects were identified as grounding targets. For each target, we meticulously designed five corresponding natural language instructions, resulting in a total of 25 unique queries per scene. As illustrated in Table \ref{tab:appendix_task} (taking the Office scene as an example), these instructions are specifically engineered to emphasize spatial relational descriptions. By utilizing various linguistic structures to describe the same target object relative to specific anchors (underlined in the table), we aim to evaluate whether the model can achieve stable and consistent grounding under diverse forms of expression. This setting also introduces substantial ambiguity, since many candidate objects may share similar semantic categories while differing only in their spatial relationships to surrounding anchors. Therefore, the task requires the model not only to recognize object-level semantics, but also to establish a fine-grained structural understanding of the 3D environment, including proximity, orientation, relative positioning, and contextual object arrangement. Such a design provides a more rigorous test of whether the model can generalize beyond simple category matching and perform reliable grounding based on compositional spatial reasoning in real-world scenes.

These real-world instructions are intended as controlled stress cases rather than exhaustive language coverage. By keeping the target object fixed while varying the expression around the same anchor, the setup separates recognition of the target category from consistency under relational paraphrase. This makes the qualitative examples complementary to the aggregate success rates reported in the main paper.

\FloatBarrier

\end{document}